\documentclass{article}
\usepackage{float}
\usepackage{colm2024_conference}

\usepackage{booktabs}
\usepackage{graphicx}
\usepackage{tabularx}
\usepackage{array}
\newcolumntype{L}{>{\raggedright\arraybackslash}X}
\usepackage{enumitem}
\usepackage{wrapfig}
\usepackage{algorithm}
\usepackage{algpseudocode}
\usepackage{natbib}
\usepackage{makecell}
\usepackage{booktabs}
\usepackage{bbm}
\usepackage{array}
\usepackage{amsmath} 
\usepackage{amssymb}
\usepackage{amsfonts}
\usepackage{multirow}
\usepackage{verbatim}
\usepackage{caption}
\usepackage{longtable}
\usepackage{supertabular}

\usepackage{CJKutf8}
\usepackage[utf8]{inputenc} % optional
\usepackage[T1]{fontenc} 
\usepackage{microtype}
\usepackage[french,vietnamese,mongolian,greek,english]{babel}

\usepackage{pifont}
\usepackage{afterpage}
\usepackage{enumitem}
\usepackage{tablefootnote}
\usepackage{xspace}
\usepackage{textcomp}
\usepackage{makecell}
\usepackage{lscape} 
\usepackage{siunitx}
\usepackage{listings}
\usepackage{xcolor}
\usepackage{adjustbox}
\definecolor{dt}{gray}{0.7}
\definecolor{tongyi-purple}{RGB}{97,92,237}
\colorlet{tongyi-purple-alpha}{tongyi-purple!38}

\usepackage{pifont}       
\usepackage{bbding}       
\usepackage{fontawesome}

\usepackage{tgpagella}
\usepackage{latexsym}
\usepackage[T1]{fontenc}
\usepackage[utf8]{inputenc}
\definecolor{mydarkblue}{rgb}{0,0.08,0.45}
\definecolor{citecolor}{HTML}{0071BC}
\usepackage{url}            
\usepackage{nicefrac}       % compact symbols for 1/2, etc.
\usepackage{changepage}
\usepackage{xargs}          % Use more than one optional parameter in a new commands
\usepackage{wrapfig,lipsum,booktabs}
\usepackage{longtable}
\usepackage{subcaption}
\usepackage{endnotes}

\usepackage{pgfplots}
\usetikzlibrary{pgfplots.groupplots}
\pgfplotsset{compat=1.3}
\usepackage{tikz}
\usetikzlibrary{patterns}

\usepackage[most]{tcolorbox}
\usepackage{fvextra}
\usepackage{graphicx}
\usepackage[capitalize,noabbrev]{cleveref}
\crefname{section}{Section}{\S\S}
\Crefname{section}{Section}{\S\S}
\crefname{table}{Table}{Tables}
\crefname{figure}{Figure}{Figures}
\crefname{algorithm}{Algorithm}{}
\crefname{equation}{eq.}{}
\crefname{appendix}{Appendix}{}
\crefformat{section}{Section #2#1#3}
\usepackage{multicol}
\usepackage{fancyvrb} 
\usepackage{tcolorbox}
\newsavebox{\myverbcontent}
\usepackage{titlesec}
\titleformat*{\section}{\large\bfseries}

\usepackage{nicematrix} % for dashline in table added by beilin
\usepackage{arydshln}
\usepackage{hyperref}

\makeatletter
\DeclareRobustCommand\onedot{\futurelet\@let@token\@onedot}
\def\@onedot{\ifx\@let@token.\else.\null\fi\xspace}

\def\eg{\emph{e.g}\onedot}

\title{Qwen-VLA: Unifying Vision-Language-Action Modeling across Tasks, Environments, and Robot Embodiments}

\author{
\bf Qwen Team}

\begin{document}

\maketitle

\begin{abstract}
Embodied intelligence is often studied through specialized models, each designed for a single scenario or task, such as manipulation and navigation, leading to fragmented capabilities and limited generalization across diverse tasks, environments, and robot embodiments. In this work, we investigate whether these heterogeneous embodied decision-making problems can be unified within a single vision-language-action model. 
We present Qwen-VLA, a unified embodied foundation model that extends Qwen’s vision-language modeling stack from perception, understanding and reasoning to continuous action and trajectory generation through a DiT-based action decoder. 
Our approach adopts a large-scale joint pretraining recipe over diverse data sources, including robotics manipulation trajectories, human egocentric demonstrations, synthetic simulation data, vision-and-language navigation data, trajectory-centric supervision, and auxiliary vision-language data.
To support multiple robot platforms within a shared model, we introduce embodiment-aware prompt conditioning, where robot-specific textual descriptions are prepended to specify the current embodiment and control convention. 
We further cast manipulation, navigation, and trajectory prediction into a unified action-and-trajectory prediction framework, enabling transferable visual grounding, spatial reasoning, and continuous action generation across robot morphologies, task families, and environments.
Experiments on manipulation, navigation, and trajectory-centric benchmarks show that Qwen-VLA supports embodied control across task families and robot embodiments, with consistent multi-task performance and out-of-distribution generalization across variations in scene layout, background, lighting, object configuration, and robot embodiment. 
As a unified generalist policy, Qwen-VLA-Instruct simultaneously achieves 97.9\% on LIBERO, 73.7\% on Simpler-WidowX, 86.1/87.2\% on RoboTwin-Easy/Hard, 69.0\% OSR on R2R, and 59.6\% SR on RxR, while further attaining 76.9\% average OOD success in real-world ALOHA experiments and 26.6\% zero-shot success rate on DOMINO dynamic manipulation.

\end{abstract}

\vspace{24pt}

  \begin{figure}[!th]                                   
      \centering
      \includegraphics[width=0.9\linewidth]{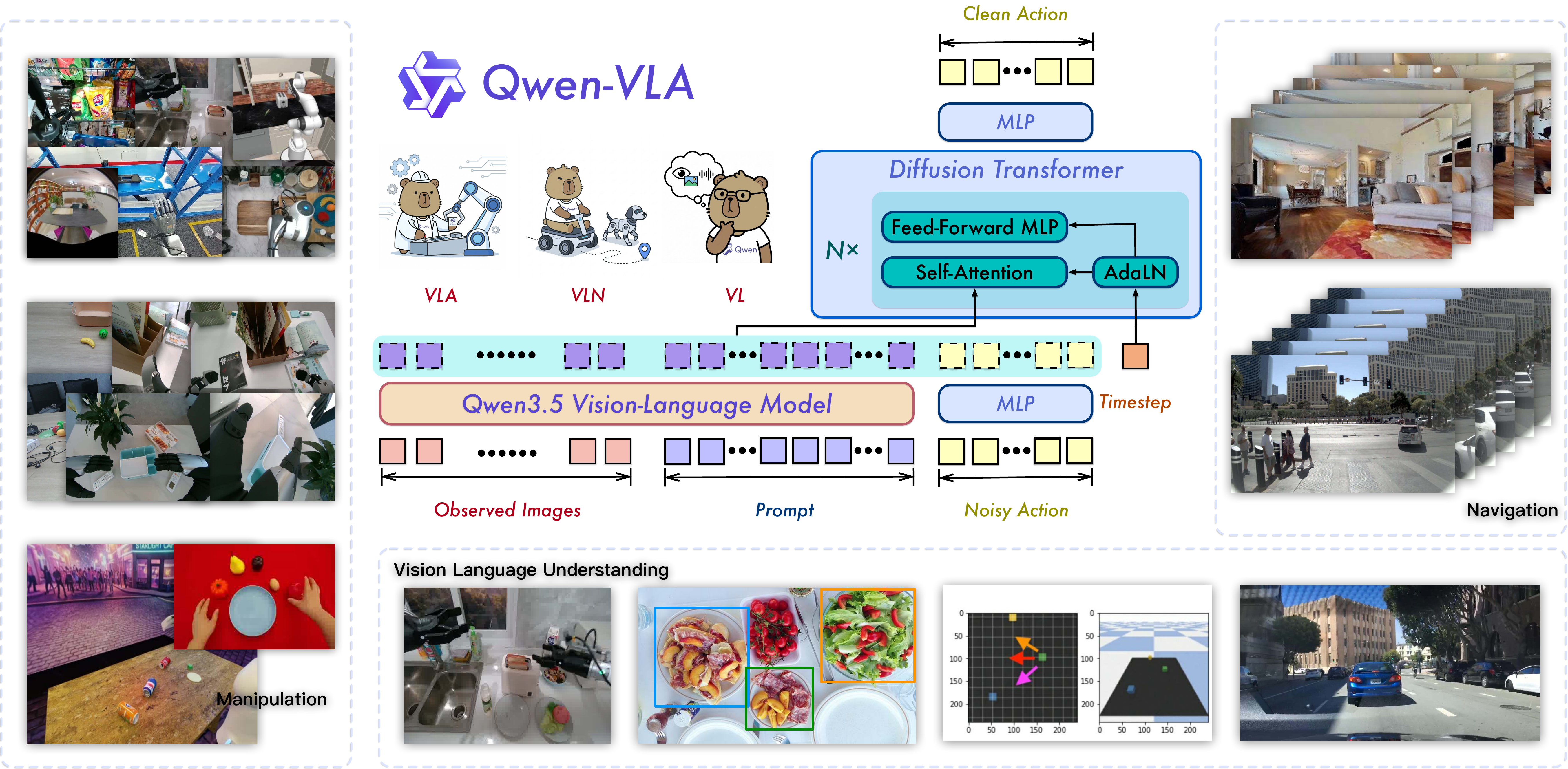}             
      \caption{Overview of Qwen-VLA, a unified embodied model trained on mixed manipulation, navigation, and vision-language understanding data to generate both robot actions and textual responses.}                                                                      
      \label{fig:overview}                                                                       
  \end{figure}

\newpage

\section{Introduction}

Embodied intelligence~\citep{zitkovich2023rt,o2024open} aims to build agents that can perceive the physical world, understand natural-language instructions, reason over spatial and temporal context, and execute actions to accomplish goals. Recent progress in vision-language models (VLMs)~\citep{Qwen2-VL,qwen2.5vl,team2025gemini,bai2025qwen3,qwen35blog} has substantially improved open-world visual understanding and language grounding, while diffusion- or flow-based policies~\citep{black2024pi0,black2025pi05,intelligence2026pi} have shown strong potential for modeling continuous, high-dimensional robot actions. However, most existing embodied systems~\citep{nvidia2025gr00t,luo2025beingh0,yang2026abotm0} remain specialized to a narrow task family, robot embodiment, or evaluation setting. Manipulation models~\citep{intelligence2026pi} are often trained for tabletop or dexterous control, and navigation models~\citep{zhang2024navid,zhang2024uni,cheng2024navila} are designed around waypoint or action prediction in indoor environments. This fragmentation limits transfer across tasks, environments, and embodiments, and makes it difficult to scale embodied learning in the same way as general-purpose vision-language pretraining.

A central challenge is that embodied decision-making problems appear heterogeneous on the surface. Robot manipulation may require predicting end-effector poses~\citep{black2024pi0,zheng2025x,zhang2026vlm4vla}, joint positions~\citep{fu2024mobile,nvidia2025gr00t}, gripper states, or dexterous-hand configurations~\citep{luo2025beingh0}; navigation may require predicting waypoints or discrete movement decisions~\citep{navfom,wang2025trackvla,wei2025streamvln}; egocentric human demonstrations may provide wrist and hand trajectories rather than robot control signals. These tasks also differ in observation format, control frequency, prediction horizon, action dimensionality, and evaluation protocol. Nevertheless, they share a common computational structure: an embodied agent must condition on visual observations, language instructions, and embodiment-specific constraints, then predict future actions or trajectories that are physically and semantically aligned with the task. This observation motivates a unified formulation for embodied modeling.
We leverage this insight to design a joint pretraining framework that absorbs manipulation, navigation, egocentric human demonstrations, and vision-language reasoning into a single vision-language-action model. The resulting model generalizes across embodiments and task settings.

In this work, we present \textbf{Qwen-VLA}, a unified vision-language-action model for heterogeneous embodied decision-making. 
Qwen-VLA is built upon Qwen3.5-4B, a natively multimodal backbone from Qwen families that has demonstrated state-of-the-art performance across a wide range of vision-language benchmarks. The backbone brings strong fine-grained visual perception, robust referential grounding, multilingual instruction following, and structured reasoning capabilities, which are critical for embodied tasks that require interpreting spatial relations, identifying referred objects, and following complex multi-step instructions.
On top of the backbone, we attach a DiT-based flow-matching action decoder that specializes in fine-grained continuous action generation. Instead of designing separate output heads or task-specific architectures for different embodiments, Qwen-VLA represents manipulation actions, navigation waypoints, and human egocentric motions in a shared action-and-trajectory prediction space. This design allows a single model to absorb supervision from diverse embodied datasets while preserving a unified inference interface.

A key component of Qwen-VLA is large-scale joint pretraining over diverse embodied and vision-language data. Our pretraining mixture includes robot manipulation trajectories, human egocentric demonstrations, synthetic simulation trajectories, vision-and-language navigation data, spatial grounding data, autonomous-driving VQA, fine-grained embodied action captions, and general vision-language data. This mixture is designed to cover both low-level motor priors and high-level semantic reasoning. Robot trajectories provide direct action supervision, egocentric human data supplies scalable real-world manipulation priors, navigation data introduces long-horizon instruction following and exploration, synthetic simulation improves controllability and out-of-distribution coverage, and vision-language data, i.e., selectively sampled for embodied-relevant capabilities such as spatial reasoning, referential grounding, and instruction following, preserves and strengthens the backbone's perception and reasoning foundations for downstream embodied tasks.

To enable cross-embodiment learning, we introduce \textbf{embodiment-aware prompt conditioning}. Each training example is prepended with a textual description of the current embodiment, including the robot platform, arm configuration, control convention, control frequency, and prediction horizon. The prompt serves as the sole interface through which the model is informed of embodiment-specific control semantics. Combined with a unified action representation, this allows the same action decoder to handle different control modes, action dimensions, and temporal horizons without changing model architecture. As a result, Qwen-VLA can learn shared visual grounding and spatial reasoning capabilities across heterogeneous embodiments while still respecting platform-specific action conventions.

Training such a model is non-trivial because the VLM backbone and action decoder enter optimization in asymmetric states: the backbone already possesses general-purpose vision-language representations from large-scale pretraining, while the DiT action decoder is randomly initialized. Naively launching multimodal joint training from this point can waste computation on vision-independent decoder learning and destabilize the pretrained representations. We therefore adopt a staged training recipe grounded in a compression view of action learning. A language instruction such as ``pick up the red cup'' together with an embodiment prompt compactly encodes the task intent in a handful of tokens, yet the corresponding action trajectory may span hundreds of high-dimensional joint-position values. Bridging this dimensionality gap is a structured decompression problem, and our first training stage, i.e., \textbf{text-to-action DiT pretraining (T2A)}, teaches the decoder to serve as a language-conditioned action decompressor before any visual input is introduced. Once this compressed action prior is in place, continued pretraining grounds it in visual observations, supervised fine-tuning specializes it for downstream tasks, and reinforcement learning optimizes closed-loop task success. This recipe separates action-prior compression, visual grounding, task specialization, and success-driven refinement into distinct stages.

We evaluate Qwen-VLA across multiple embodied settings, including robot manipulation, vision-and-language navigation, out-of-distribution generalization, and cross-embodiment transfer. The evaluation is designed not only to measure in-domain task success, but also to test whether a unified model can generalize under changes in object layout, lighting, background, language instruction, sensor noise, and robot embodiment. Empirically, our results show that large-scale joint pretraining over heterogeneous embodied data yields strong multi-task performance and improves robustness beyond narrow specialist training. These findings support the view that manipulation, navigation, and trajectory-centric embodied tasks can be treated as different manifestations of a shared action-and-trajectory prediction problem.

Our contributions are summarized as follows:

\begin{itemize}
      \item We introduce \textbf{Qwen-VLA}, a unified vision-language-action model that formulates manipulation, navigation, and egocentric action modeling within a shared action-and-trajectory space. Built on the Qwen3.5-4B vision-language backbone with a DiT-based flow-matching policy head, Qwen-VLA supports embodied control across multiple robot platforms and task families.

      \item We construct a large-scale joint pretraining mixture over heterogeneous data sources, including manipulation trajectories from multiple robots, egocentric human demonstrations, synthetic simulation, navigation, and curated vision-language data. We further propose embodiment-aware prompt conditioning that unifies diverse robot platforms, control conventions, and prediction horizons in one model without separate per-embodiment policies.

      \item We design a progressive training recipe that includes action pretraining, multimodal continued pretraining, supervised fine-tuning, and reinforcement learning to build a strong policy model that bridges the gap between discrete vision-language tokens and continuous action trajectories, improving both training stability and downstream transfer.
      
      \item We evaluate Qwen-VLA across manipulation, navigation, out-of-distribution robustness, and cross-embodiment generalization benchmarks. The results show that co-training and progressive learning improve multi-task performance under scene, object, lighting, and embodiment shifts.
  \end{itemize}

\section{Unified Embodied Model}
\subsection{Problem Formulation}

We study a broad family of embodied decision-making tasks, including robot manipulation, vision-and-language navigation, trajectory prediction, and human egocentric action modeling. We propose to address these tasks within a unified embodied model, motivated by their shared computational structure: despite differences in output format and evaluation protocol, all tasks require  an agent to ground language in visual observations, reason over spatial and temporal context, and predict future actions or trajectories.

We formulate all tasks in a unified conditional prediction framework. At time step $t$, the model receives a visual context $o_{t}$, a language instruction $x$, an embodiment description $e$, and an optional task identifier $z$. Here, $o_{t}$ may consist of one or multiple image frames, video observations, or history windows; $x$ specifies the task instruction; $e$ is a textual prompt describing the current robot platform and control convention; and $z$ identifies the task family when needed. The model is trained to predict a target sequence $y_{t:t+H-1}$ over a prediction horizon $H$:
\[
p_\theta(y_{t:t+H-1} \mid o_{t}, x, e, z).
\]

The target sequence $y_{t:t+H-1}$ is task-dependent but represented in a unified action-and-trajectory space. For manipulation tasks, it corresponds to future robot actions like end-effector positions; for navigation tasks, it represents navigation decisions or waypoints; for trajectory-centric tasks such as autonomous driving or motion forecasting, it denotes the precise future spatial trajectory of the agent or surrounding entities in continuous coordinate space; and for egocentric embodied data, it captures human body or hand motion trajectories represented in structured pose spaces such as MANO~\citep{MANO} or skeletal joint sequences. This unified formulation enables joint optimization across heterogeneous embodied datasets within a single model, facilitating transferable visual grounding, spatial reasoning, and action generation across task families. The formulation is also extensible along both the input and output axes: on the input side, augmenting the conditioning context $o_{t}$ with episodic memory or persistent state would enable long-horizon planning and failure recovery; on the output side, co-predicting future visual states alongside actions would unify action generation with world modeling, allowing the agent to anticipate the consequences of its actions. 
\subsection{Model Architecture}
\label{sec:architecture}

Our model comprises a vision-language backbone for high-level understanding and reasoning and a flow-matching action expert for fine-grained action generation.

\paragraph{Vision-language backbone.}
We adopt Qwen3.5~\citep{qwen35blog} as the backbone. Qwen3.5 is a natively multimodal model trained with early vision-language fusion: visual tokens produced by a ViT with spatial merging are interleaved directly into the text token stream, enabling unified processing of images,
videos, and language within a single transformer.
Its hybrid attention design combines gated linear attention in the majority of layers with grouped-query softmax attention at regular intervals, allowing
efficient encoding of long multimodal sequences while retaining full-precision global reasoning where needed.

\paragraph{Action expert.}
We attach a single-stream DiT-style~\citep{esser2024scaling} flow matching policy as an action expert for predicting precise actions across both robot and human embodied data~\citep{janner2022planning,chi2023diffusion,adaptdiffuser,black2024pi0}.
The action expert concatenates VLM hidden states with a noisy action chunk into one sequence and processes them through joint self-attention with AdaLN timestep
conditioning~\citep{peebles2023scalable} and multi-section RoPE aligned with the backbone.
This decoupled design lets the action expert specialize in fine-grained action generation and naturally handle the multimodality and high-frequency dynamics of
embodied action distributions, while preserving the backbone's pretrained capabilities.
The expert is trained with a flow-matching objective~\citep{lipman2023flow} and produces action sequences via a few Euler integration steps at inference,
enabling low-latency real-time control.
In total, our action expert contains approximately 1.15B parameters: 16 DiT blocks account for the bulk (70.8M each, 1.13B combined), with the remaining parameters distributed among action projection MLPs that
map between the raw action dimension and the DiT latent space (4.9M), a linear layer that transforms VLM hidden states to the DiT channel dimension (3.9M), timestep embedding (2.8M), and output AdaLN modulation
(4.7M).

\subsection{Embodiment-aware Prompt Conditioning}
\label{sec:Prompt-formulation}
\label{sec:embodiment}

To support multiple robot embodiments within one shared model, we prepend a
robot-specific textual prompt to each training example that describes the
current platform, arm configuration, and control convention.
The prompt follows the template:

\begin{quote}
\texttt{The robot is \{robot\_tag\} with \{single arm / dual arms\}[, waist][, and mobile base].
The control frequency is \{FPS\} Hz.
Please predict the next \{chunk\_size\} control actions to execute the following task: \{ori\_instruction\}.}
\end{quote}

\noindent The robot tag and optional modifiers (\emph{waist}, \emph{mobile base}) are set per
embodiment; \texttt{FPS} and \texttt{chunk\_size} reflect the dataset's original control
frequency and prediction horizon.
Table~\ref{tab:robots} in \cref{sec:data} summarises the representative robot
platforms covered by our pretraining corpus together with their arm
configurations and action types.

\subsection{Unified Action and Trajectory Representation}
\label{sec:unified-representation}
% We unify all continuous control signals into a single tensor format consumed by the DiT-based action expert, regardless of embodiment or task family.
We unify the tensor interface and masking scheme, but do not force all embodiments into a single physical action semantic space. Each dataset preserves its native control convention, specified through the embodiment prompt and dataset-specific normalization.
Concretely, each training sample contributes a target tensor $\mathbf{Y} \in \mathbb{R}^{H \times K}$, where $H$ is a fixed prediction horizon and $K$ is a fixed channel dimension shared across all control modes.

% We unify the tensor interface and masking scheme, but do not force all embodiments into a single physical action semantic space. Each dataset preserves its native control convention, specified through the embodiment prompt and dataset-specific normalization.

\paragraph{Control signal types.}
We cover two families of continuous control signals.
\textit{Manipulation} signals include delta end-effector position ($\Delta x, \Delta y, \Delta z$), end-effector rotation expressed as Euler angles or quaternions, absolute joint positions, gripper aperture, and dexterous-hand joint angles.
\textit{Navigation trajectory} signals follow the VLN convention and are represented as $(\Delta x, \Delta y, \Delta\theta)$ per waypoint, encoding relative displacement and heading change in the ground plane.
Despite their different physical semantics, both families are sequences of real-valued vectors predicted over a horizon, and are therefore treated identically by the action expert.

\paragraph{Channel layout.}
\label{sec:unified_rep:channel_mask}
A given control mode uses $c \leq K$ channels.
These $c$ task-relevant values are placed in the \emph{leading} $c$ dimensions of $\mathbf{Y}$, and the remaining $K - c$ dimensions are zero-padded.
A per-channel binary mask $\mathbf{M} \in \{0,1\}^{H \times K}$ records which channels carry valid signals: $M_{h,k} = 1$ if and only if channel $k < c$ and time step $h$ falls within the task's chunk length $H_{\text{task}} \leq H$.
This scheme requires no embodiment-specific output heads; a single set of DiT parameters handles all control modes, with the mask preventing padded entries from influencing the gradient.

\paragraph{Task-aware conditioning.}
Each training sample is prefixed with the embodiment-aware prompt described in \cref{sec:embodiment}, which specifies the robot platform, arm configuration, control frequency, and prediction horizon.
For VLN samples the prompt analogously states the navigation convention and waypoint horizon.
These prompt tokens are processed by the VLM backbone and their hidden states are concatenated with the noisy action chunk as input to the DiT, so the action expert is always conditioned on the precise control specification of the current sample without any architectural changes.

\subsection{Training Objectives}
We train the entire model end-to-end with a weighted sum of two objectives that cover continuous action generation and vision-language understanding.

\paragraph{Flow-matching action loss.}
For all samples with continuous control targets (manipulation, VLN trajectory waypoints, and human egocentric data after action alignment), we supervise the action expert with a conditional flow-matching objective~\citep{lipman2023flow}.
Given a clean target $\mathbf{Y}_0 \in \mathbb{R}^{H \times K}$ and noise $\mathbf{Y}_1 \sim \mathcal{N}(0, \mathbf{I})$, we form the linear interpolant $\mathbf{Y}_\tau = (1-\tau)\mathbf{Y}_0 + \tau \mathbf{Y}_1$ with $\tau \in [0,1]$, and train the expert $v_\theta$ to predict the conditional velocity field.

To avoid the gradient being dominated by padding, we apply a \emph{per-channel, per-step} loss with two levels of averaging.
For a sample whose control mode activates $c$ channels and $H_{\text{task}}$ time steps, let $\mathbf{M} \in \{0,1\}^{H \times K}$ be the validity mask defined in \cref{sec:unified_rep:channel_mask}.
We first compute the mean squared error for each active channel $k < c$:
\begin{equation}
\ell_k = \frac{
    \displaystyle\sum_{h=1}^{H} M_{h,k}
    \left\| \bigl(v_\theta(\mathbf{Y}_\tau, \tau \mid o_{1:t}, x, e, z) - (\mathbf{Y}_1 - \mathbf{Y}_0)\bigr)_{h,k} \right\|_2^2
}{
    \displaystyle\sum_{h=1}^{H} M_{h,k}
},
\end{equation}
and then average uniformly over the $c$ active channels:
\begin{equation}
\mathcal{L}_{\text{act}} = \mathbb{E}_{\tau,\mathbf{Y}_0,\mathbf{Y}_1} \left[ \frac{1}{c} \sum_{k=0}^{c-1} \ell_k \right].
\end{equation}
This two-level averaging ensures that each control dimension contributes equally to the gradient regardless of how many channels a given embodiment uses, and that padded positions are fully excluded.
At inference, action chunks are produced by a few Euler integration steps from $\tau=1$ to $\tau=0$.

\paragraph{Vision-language loss.}
To preserve and strengthen the multimodal capabilities of the backbone, we keep a standard next-token prediction loss on auxiliary vision-language data, fine-grained embodied action captions, autonomous driving VQA, and general VL pretraining corpora:
\begin{equation}
\mathcal{L}_{\text{vl}} = - \sum_{i} \log p_\theta(w_i \mid w_{<i}, o_{1:t}),
\end{equation}
where $w_i$ are text tokens. This objective stabilizes language grounding under heavy embodied co-training and prevents catastrophic forgetting of perception and reasoning skills.

\paragraph{Joint objective.}
The overall training loss is a weighted combination
\begin{equation}
\mathcal{L} = \lambda_{\text{act}} \mathcal{L}_{\text{act}} + \lambda_{\text{vl}} \mathcal{L}_{\text{vl}},
\end{equation}
where the weights $\lambda_{\text{act}}, \lambda_{\text{vl}}$ are tuned to balance the gradient magnitudes of the two objectives. Within each mini-batch we mix samples from all task families according to a fixed sampling ratio, so that every optimization step jointly updates the backbone and the action expert with manipulation, VLN trajectory, and vision-language signals.

\section{Large-Scale Joint Pretraining}

\subsection{Training Recipe} 
\label{sec:training-recipe}

  \begin{figure}[t]                                   
      \centering
      \includegraphics[width=\linewidth]{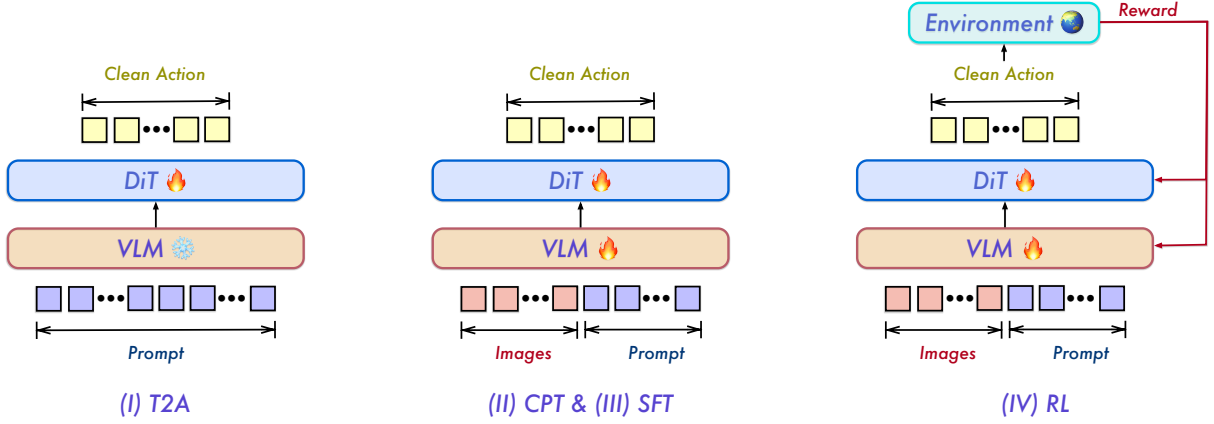}             
      \caption{Training recipe of Qwen-VLA. Stage~I (T2A) trains the DiT action decoder to reconstruct actions from text alone, building a structured action prior without visual input. Stage~II (CPT) unfreezes both modules to ground this prior in visual observations. Stage~III (SFT) branches into multi-task and real-robot tracks, and Stage~IV (RL) optimizes closed-loop task success via environment rewards.}                                                                      
      \label{fig:recipe}                                         \vspace{-5pt}                              
  \end{figure}

A usable VLA model requires co-training a cognitive backbone and a motor
decoder, a division of labor analogous to the complementary roles of the
cerebrum and cerebellum
in biological motor control.
In practice, however, these two modules enter training in deeply asymmetric
states: the VLM backbone is already strongly pretrained, while the DiT action
decoder is randomly initialized.
Naive joint training from this starting point is inefficient and unstable.
The decoder has to learn several things at once: the shape of the action
distribution, how to condition on language and embodiment, the flow-matching
dynamics of its own parameterization, and how to ground actions in vision.
At the same time, every step pays the cost of encoding images.
Meanwhile, noisy gradients from a fresh decoder may perturb the pretrained
backbone before the decoder has learned useful action structure.

Our staged recipe is motivated by a \emph{compression} perspective on action
learning.
Raw action trajectories are dense, high-frequency, and
embodiment-dependent: a single manipulation episode may contain thousands of
joint-position values spread across dozens of degrees of freedom.
Yet the underlying task intent is compactly captured by a language instruction
(``pick up the red cup'') and an embodiment prompt specifying the robot platform
and control convention. This description fits in a handful of tokens.
There exists a vast dimensionality gap between this compressed task description
and the full action signal; bridging it is a structured \emph{decompression}
problem.

We cast T2A as learning this decompression map.
By withholding images and training the DiT solely on language-conditioned
action prediction, we force the decoder to acquire a structured prior over the
action space that is indexed entirely by language.
This is more than a warm-start: the decoder learns how different linguistic descriptions select different regions of the
action distribution, how embodiment prompts modulate the same task intent into
platform-specific motor programs, and the temporal coherence and compositionality of full action trajectories
at the sequence level, using only the compressed description.
With this language-indexed action prior in place, subsequent multimodal
training can focus its capacity on grounding the prior in concrete visual
observations rather than learning action generation from scratch.

Based on this principle, we adopt a four-stage training recipe on top of a
pretrained Qwen3.5 VLM backbone:
(I)~\textbf{text-to-action DiT pretraining (T2A)},
(II)~\textbf{continued pretraining (CPT)},
(III)~\textbf{supervised fine-tuning (SFT)} in two parallel branches, and
(IV)~\textbf{reinforcement learning (RL)}.
Each stage is defined by the gap it closes in the one before it.

\paragraph{Stage~I: Text-to-action DiT pretraining (T2A).}
We freeze the VLM and train only the DiT, conditioning it on text and the
  embodiment prompt~$e$ (\cref{sec:embodiment}), but deliberately withholding
  images.
  The decoder thus operates as a pure language-to-action decompressor: it
  must reconstruct high-dimensional action distributions from the compact
  linguistic encoding alone, without any visual shortcut.
  In doing so, T2A installs a structured action prior in which language
  selects the region of action space, the embodiment prompt specifies the
  platform-specific motor parameterisation, and the flow-matching dynamics
  govern the generative process before any image is introduced.

  \paragraph{Stage~II: Continued pretraining (CPT).}
  With the decoder warm-started, CPT focuses its multimodal capacity on exactly
  the problem that T2A could not address: grounding actions in visual
  observations and adapting the backbone to embodied perception.
  We unfreeze both modules and train on the heterogeneous mixture of
  \cref{sec:data}, which deliberately combines simulation and real-robot
  trajectories so that the resulting checkpoint has seen
  both domains.
  This property is important for the post-training stages described below:
  SFT and RL can specialise toward either domain.

  \paragraph{Stage~III: Supervised fine-tuning (SFT).}
  While CPT provides broad cross-embodiment and cross-task coverage,
  downstream deployment benefits from aligning the model with curated,
  high-quality demonstrations drawn from the target task distribution.
  We branch SFT into two parallel tracks from the CPT checkpoint.
  The first track performs multi-task SFT that jointly fine-tunes the model on heterogeneous tasks, including visual
  question answering, spatial grounding, manipulation, and navigation, under
  embodiment-balanced and task-balanced sampling.
  The second track fine-tunes from the CPT checkpoint on in-house
  teleoperation data for real-world robot deployment, testing whether CPT's
  cross-domain priors transfer to physical hardware.

  \paragraph{Stage~IV: Reinforcement learning (RL).}
  SFT optimises a likelihood over demonstrations; what ultimately matters is
  closed-loop task success, a property of executed trajectories that no
  imitation objective can directly optimise.
  Starting from the multi-task SFT checkpoint, RL
  fine-tunes the policy with sparse binary success rewards collected
  exclusively in a single simulation environment (SimplerEnv), producing the
  final model Qwen-VLA-Instruct.
  This deliberately narrow RL setup tests whether task-success improvements
  obtained in one simulated environment can transfer to other
  out-of-distribution environments.
  
\subsection{Pretraining Data}
\label{sec:data}

The quality and diversity of the pretraining corpus directly determine how well
the cognitive backbone and motor decoder co-adapt across embodiments and task
families.
We construct a large and heterogeneous pretraining mixture to equip the model
with broad embodied perception, spatial reasoning, and action generation
capabilities.
The mixture spans five data families: robot manipulation trajectories, human
egocentric demonstrations, synthetic simulation data, navigation and
trajectory-centric data, and auxiliary vision-language data.
Table~\ref{tab:data} summarizes the composition and sampling weights of each
source.

\begin{table}[h]
\centering
\caption{Pretraining data mixture composition.}
\label{tab:data}
\begin{tabular}{lc}
\toprule
\textbf{Data Source} & \textbf{Proportion (\%)} \\
\midrule
Robot Manipulation Trajectories      & 74.2 \\
Human Egocentric Trajectories                & 6.0  \\
Navigation Trajectories                     & 7.5  \\
Synthetic Simulation Trajectories (ours)            & 3.7  \\
General Vision-Language Data         & 3.4  \\
Spatial Grounding (2D)               & 2.5  \\
Autonomous Driving VQA               & 2.4  \\
Fine-Grained Embodied Action Caption & 0.2  \\
\midrule
\textbf{Total}                       & \textbf{100.0} \\
\bottomrule
\end{tabular}
\end{table}

\subsubsection{Robotics Manipulation Trajectories}

Real and simulated robot manipulation trajectories form the core of our
pretraining corpus, accounting for approximately 74.2\% of the total
pretraining mixture. The data spans tabletop manipulation, mobile manipulation,
bimanual tasks, and dexterous hand control across a diverse set of robot
embodiments.

\paragraph{Public datasets.}

To improve data scale, diversity, and cross-embodiment generalization, we train
on a broad mixture of publicly available real-robot datasets, including
RobotSet~\citep{RoboSet}, Galaxea~\citep{galaxea_g0},
AgiBot World~\citep{agibot_world_colosseo}, RoboCOIN~\citep{robocoin},
RoboMIND V1/V2~\citep{robomind_v1,robomind_v2}, RDT-1B~\citep{RDT-1B},
DROID~\citep{droid}, BridgeData V2~\citep{bridgev2},
RH20T~\citep{RH20T}, RT-1~\citep{rt1}, and BC-Z~\citep{bc_z}.
These datasets cover diverse robotic settings, including tabletop
manipulation, mobile manipulation, bimanual manipulation, dexterous hand
control, and in-the-wild task execution, amounting to over 10{,}000 hours of
interaction data across heterogeneous embodiments and task scenarios. This
large-scale mixture provides complementary supervision over robot morphologies,
camera viewpoints, object categories, backgrounds, language instructions, and
action distributions, thereby reducing overfitting to any single embodiment or
environment and improving robustness under distribution shifts.

In addition to real-robot data, we include simulation-based manipulation
trajectories from InternData-A1~\citep{interndata-a1} and
GR00T-X-Embodiment-Sim~\citep{nvidia2025gr00t}, which are generated through motion
planning in diverse virtual environments. These simulated trajectories augment
scene and object diversity, especially for long-tail object configurations and
task layouts. Before training, all real and simulated datasets are curated and
converted into a unified observation-action format while preserving their
original task semantics, enabling scalable training over heterogeneous robot
experience.
\paragraph{Proprietary datasets.}
We supplement with over 1{,}000 hours of in-house collected real-robot trajectories
covering a variety of manipulation tasks and robot platforms, together
contributing approximately 20\% of the total pretraining mixture.
In addition, we generate over 8M synthetic simulation trajectories using our
scalable simulation pipeline illustrated in \cref{subsubsection:SyntheticSimulationData}, contributing a further 3.7\%
of the mixture and substantially expanding the diversity of scenes, objects,
and task configurations beyond what real-robot teleoperation alone can cover.

\paragraph{Embodiment-aware prompt conditioning.}
As described in \cref{sec:embodiment}, each training example is prepended with
a robot-specific prompt specifying the current platform, arm configuration, and
control convention.
Table~\ref{tab:robots} lists the representative embodiments in our pretraining
corpus with their arm configurations and action types.

\begin{table}[h]
\centering
\caption{Representative robot embodiments in the pretraining corpus.
``EEF'' = end-effector pose; ``Joint'' = joint angles;
``$\Delta$'' = delta (relative) commands; ``Abs'' = absolute commands;
``G'' = gripper; ``DH'' = dexterous hand.}
\label{tab:robots}
\setlength{\tabcolsep}{4pt}
\begin{tabular}{llll}
\toprule
\textbf{Robot} & \textbf{Arms} & \textbf{Action type} \\
\midrule
WidowX            & Single & $\Delta$EEF + G \\
Google Robot      & Single & $\Delta$EEF + G \\
Franka Panda      & Single / Dual & $\Delta$EEF + G;\; Abs Joint + G \\
ARX5              & Dual   & $\Delta$EEF + G \\
Fourier GR-1      & Dual   & $\Delta$EEF + G \\
Mobile ALOHA      & Dual   & $\Delta$EEF + G;\; Abs Joint + G \\
AgiBot A2-D       & Dual   & Abs Joint + G;\; Abs Joint + DH \\
Galaxea R1        & Dual   & Abs Joint + G \\
AIRBOT MMK2       & Dual   & Abs Joint + DH \\
TienKung          & Dual   & Abs Joint + G;\; Abs Joint + DH \\
Real Human        & Dual   & $\Delta$EEF (from MANO) \\
\bottomrule
\end{tabular}
\end{table}

\paragraph{Action representation.}
Different datasets adopt different action conventions: some provide absolute
end-effector poses in Cartesian space~\citep{xie2026unifyrobotactionscamera}, others provide delta end-effector
commands, and others use absolute or relative joint-space control signals.
We preserve each dataset's original action format rather than converting to a
shared representation, relying on embodiment-aware prompt conditioning
to inform the model of the current control convention.

All action dimensions are normalized per dataset using quantile statistics.
For each action dimension $d$ in dataset $k$, we compute the 1st and 99th
percentiles $q^k_{01}$ and $q^k_{99}$ over all trajectories and apply a
linear mapping:
\begin{equation}
    \tilde{a}_d = 2 \cdot \frac{a_d - q^k_{01}}{q^k_{99} - q^k_{01}} - 1,
\end{equation}
clipping the result to $[-1, 1]$. This per-dataset quantile normalization
removes scale differences across embodiments and action spaces while preserving
relative motion structure within each source.

\paragraph{Language instructions.}
Language instructions are sourced from a combination of original dataset
annotations and model-generated captions provided alongside the data.
All instructions undergo a quality filtering and consistency checking pipeline:
trajectories whose language annotations are inconsistent with the observed
motion are discarded, ensuring reliable language--action alignment across
diverse sources.
Beyond action-labelled data, we incorporate fine-grained embodied video--caption
pairs (described in \cref{sec:vl-data}) as auxiliary vision-language supervision.
This dense language signal bridges the gap between coarse task labels and rich
semantic descriptions of robot behaviour, enabling the model to generalise to a
broader range of natural-language instructions at inference time.

\paragraph{Camera view representation.}
Many robot datasets provide observations from multiple camera viewpoints
simultaneously, \eg a head-mounted ego-centric camera alongside one or two
wrist-mounted cameras.
To make the view provenance explicit to the model, we wrap each image in the
token stream with a pair of view-specific boundary tokens:
\begin{center}
\texttt{<|}\textit{tag}\texttt{\_start|>}\ \textlangle image\textrangle\ \texttt{<|}\textit{tag}\texttt{\_end|>}
\end{center}
where \textit{tag} identifies the camera source, \eg \texttt{ego},
\texttt{cam\_left\_wrist}, or \texttt{cam\_right\_wrist}.
The explicit view labels allow the VLM backbone to form view-aware
representations and let the action expert attend selectively to each
viewpoint when producing actions, without requiring any architectural
changes or additional input channels.

\paragraph{Data cleaning.}
We apply a multi-stage filtering pipeline to remove trajectories with corrupted
or missing frames, near-zero-variance action sequences indicative of static
recordings, and episodes with anomalous lengths. For datasets without explicit
action labels, pseudo-actions are recovered via finite differences on
proprioceptive state sequences.

\subsubsection{Egocentric Human Data}
\label{sec:ego-data}

Compared with teleoperated robot trajectories, egocentric human demonstrations
offer a more abundant and scalable source of real-world manipulation
experience.
Humans interact with diverse objects in open-world environments every day,
naturally producing dexterous manipulation behaviours that span a much broader
range of scenes, objects, and task semantics than robot teleoperation.
Recent studies demonstrate that training on large-scale egocentric human videos endows vision-language-action models with richer manipulation priors and improves generalization to downstream robot tasks~\citep{kareer2025egomimic,luo2025beingh0,li2025vitra,luo2026beingh05,zheng2026egoscale,hu2026finevlafinegrainedinstructionalignment}.
Motivated by these findings, we incorporate a diverse collection of egocentric
human manipulation datasets (6.0\% of the total pretraining mixture) to
provide the model with broad manipulation priors that complement the robot
trajectory data.

\paragraph{Data sources.}
Our egocentric corpus is drawn from four publicly available datasets, each
contributing complementary coverage in terms of task diversity, scene variety,
and annotation granularity. %(Table~\ref{tab:ego-data}).
\textbf{(1)}~We adopt the Ego4D~\citep{grauman2022ego4d} and
EPIC-KITCHENS~\citep{damen2022rescaling} subsets processed by
VITRA~\citep{li2025vitra}, which applies an automated pipeline to segment egocentric human videos into atomic manipulation trajectories and generate fine-grained language annotations, together with framewise 3D hand
and camera motion trajectories.
\textbf{(2)}~EgoDex~\citep{hoque2026egodex} is a large-scale dexterous manipulation dataset captured with Apple Vision Pro, comprising 829 hours of egocentric video with paired 3D hand and finger tracking across 194 diverse
tabletop tasks.
\textbf{(3)}~EgoVerse~\citep{punamiya2026egoverse} is a collaborative egocentric demonstration platform spanning over 1{,}300 hours across 1{,}965 tasks and 240 scenes, with standardized formats and manipulation-relevant annotations.
\textbf{(4)}~Xperience~\citep{ropedia2026xperience} is a large-scale egocentric multimodal dataset providing synchronized first-person recordings with depth, hand and body motion capture, and hierarchical language annotations.

\paragraph{Action representation.}
The model receives an egocentric image at an arbitrary time step together with a language instruction and predicts an action chunk comprising future bimanual wrist motion and hand articulation over a fixed prediction horizon. 
For each hand, the wrist motion is represented as the SE(3) transformation of the wrist coordinate frame at a future time step relative to the initial frame. During training, this relative action is represented with a translation vector and an axis-angle rotation, yielding 6 wrist action dimensions per hand.
To represent hand articulation compactly, we perform principal component analysis (PCA) on the 45-dimensional axis-angle joint pose of the hand across all human datasets and retain the weights of the first 10 principal components. These low-dimensional coefficients, dubbed \emph{eigengrasps}~\citep{ciocarlie2007dexterous, yuan2024crossdex}, capture the dominant modes of human hand-pose variation while discarding per-joint redundancy. The model predicts relative wrist actions and 10 eigengrasp coefficients per hand, yielding a total of 32 action dimensions per time step for egocentric human data.

\subsubsection{Synthetic Simulation Data}
\label{subsubsection:SyntheticSimulationData}
To improve the coverage, controllability, and robustness of embodied supervision, we construct a large-scale synthetic simulation data pipeline with two complementary components: (1) \emph{vision-language-action data}, where the model predicts actions from both task instructions and image observations, and (2) \emph{language-action data}, where the model predicts actions from language alone. These two sources of supervision play different but synergistic roles. The text-only component encourages the model to acquire high-level task abstractions and language-action regularities without depending on visual appearance, while the vision-conditioned component grounds those abstractions in realistic perceptual observations, scene variations, and embodied interaction dynamics. We developed our pipeline based on IsaacLab \citep{mittal2025isaaclab} for our simulation environment and cuRobo \citep{sundaralingam2023curoboparallelizedcollisionfreeminimumjerk} for collision-avoidance motion planning.

\paragraph{Vision-language-action data.}
For the vision-conditioned synthetic data, we use an internal early version of \textsc{RoboInF}~\citep{roboinf} to generate standard VLA-style supervision, where the model receives a language instruction and image observations as input and predicts robot actions as output. \textsc{RoboInF} constructs manipulation scenes, generates scene-conditioned tasks, synthesizes executable success checks, produces robot motion programs through simulator feedback, and rolls out successful trajectories under domain randomization. Since the full technical details of the pipeline are provided in our \textsc{RoboInF} blog post, we only summarize the dataset composition and diversity here.

In this release, we focus on the random-placement scene generation setting for simplicity. We construct 20 tabletop scenes, each augmented with 10 different object initial-pose configurations, resulting in 200 base scene configurations. These scenes contain diverse everyday objects arranged in physically valid tabletop layouts, providing varied visual and spatial contexts for manipulation.

\begin{figure}[t]
    \centering
    \includegraphics[width=\linewidth]{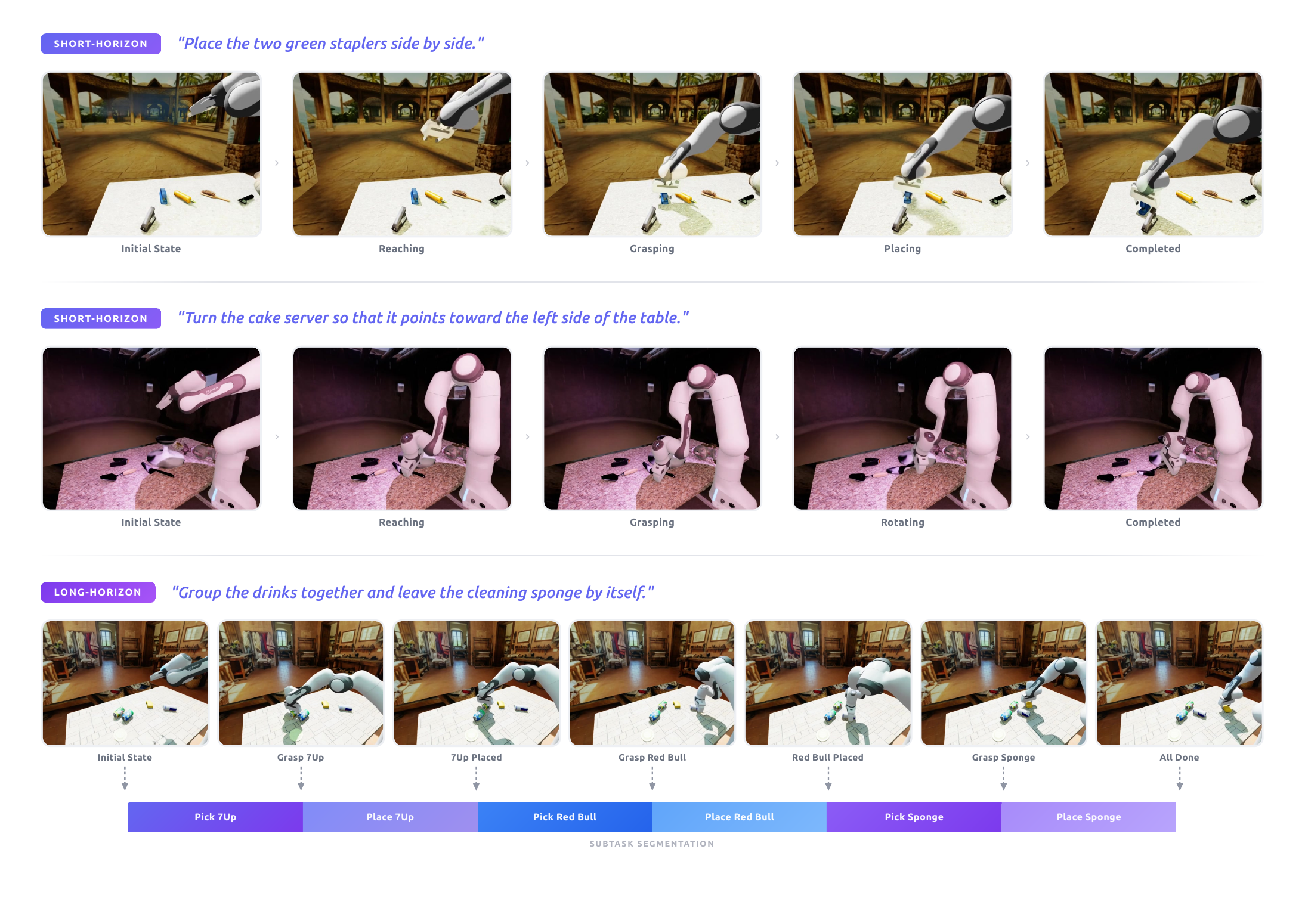}
    \caption{
    Examples of data generated through \textsc{RoboInF}. 
    The top row shows a short-horizon task, \textit{``Place the two green staplers side by side,''} which consists of a compact sequence of reaching, grasping, transporting, and placing.
    The bottom row shows a long-horizon task, \textit{``Group the drinks together and leave the cleaning sponge by itself,''} which requires multiple object manipulations and can be decomposed into subtask segments such as picking and placing each drink.
    }
    \label{fig:synthetic-sim-examples}
\end{figure}

On top of these scenes, we generate 450 manipulation tasks spanning both short- and long-horizon behaviors. Short-horizon tasks typically involve a small number of primitive manipulation steps, such as \textit{``Place the two green staplers side by side''}. Long-horizon tasks require multiple sequential interactions or more compositional reasoning, such as \textit{``Group the drinks together and leave the cleaning sponge by itself''}, where the robot needs to manipulate several objects in sequence. As illustrated in Figure~\ref{fig:synthetic-sim-examples}, this task mixture exposes the model to both atomic manipulation skills and higher-level instruction-following patterns over extended temporal horizons.

For each task, we generate 300 successful trajectories with different environment and execution augmentations. During rollout, we randomize visual, geometric, and control-related factors, including lighting, camera poses, backgrounds, table textures, robot initial states, object initial poses, and controller dynamics such as stiffness and damping. For visual randomization, we sample from a large pool of approximately 3K backgrounds and 1K table textures. For other factors, such as camera pose, robot state, lighting, and controller parameters, we define reasonable ranges manually and sample variations within those ranges. These augmentations are designed to improve robustness to distribution shifts in appearance, viewpoint, layout, and low-level execution dynamics.

A useful property of this generated data is that the trajectories are produced from motion-planning programs that are naturally decomposed into intermediate stages. We therefore additionally segment each full trajectory into several subtask trajectories, as shown in Figure~\ref{fig:synthetic-sim-examples}. This provides supervision at multiple temporal granularities: full trajectories teach the model to follow high-level instructions over longer horizons, while subtask trajectories expose the model to shorter atomic behaviors and intermediate goals. We expect this mixture to help the model better connect high-level language instructions with the lower-level action segments needed to complete them.

Overall, the vision-conditioned synthetic data contains 359,848 full successful trajectories including subtask segments. The resulting dataset provides a large-scale source of diverse, controllable, and automatically verified VLA supervision, complementing real-world demonstrations and improving the model's generalization across task types, visual conditions, and execution variations.

\paragraph{Language-action data.}
We additionally construct a text-only action dataset as a complementary source of supervision. The goal of this component is to expose the model to a broad space of manipulation intents and action patterns before introducing visual observations, thereby encouraging generalization in the language-action joint space. Concretely, we brainstorm a diverse set of atomic and compositional manipulation tasks that can be executed by the robot in simulation, and generate corresponding trajectories using robot-state-centric pipelines without image inputs. This data primarily serves as a form of semantic and behavioral pretraining, helping the model capture common task structures such as moving, grouping, separating, aligning, rotating, stacking, and reordering objects from language instructions alone.

Specifically, we define six task template families that span the core single-arm manipulation primitives: pick-and-place, linear pushing, linear pulling, rotation with repositioning, rotation toward a viewpoint direction, and positional swapping of two objects. Each template encodes a canonical motion structure while randomizing the spatial parameters and natural language description across instances. Instructions are procedurally composed from template-specific vocabularies of verbs, prepositions, object names, and spatial references to produce varied descriptions of the same underlying behavior (e.g., ``Place the apple into the container,'' ``Move the sponge toward the target location''). This design provides precise control over task coverage and difficulty while remaining simple enough to scale to millions of trajectories.

To ensure broad coverage of embodiment-specific kinematics and workspace geometry, we instantiate all six task templates across six single-arm robot configurations (Franka Panda, UR10e, UR5e, Kinova Gen3, TM12, and xArm7) spanning different degrees of freedom, arm morphologies, and reachable workspaces. For each robot--task pair, we generate approximately 200k trajectories, yielding roughly 7.2M trajectories and over 14,000 hours of simulated robot trajectory data in total. Trajectory synthesis proceeds without physics simulation or scene rendering: for each instance, we sample end-effector goal poses within the robot's reachable workspace bounds, randomizing positions, orientations, and inter-pose distances, and compute collision-free joint-space paths using a GPU-accelerated batch motion planner. The generation pipeline is parallelized across multiple GPUs, with each worker independently sampling target configurations, planning motions, and writing successful trajectories. Only kinematically valid trajectories for which the planner produces feasible, collision-free solutions across all required motion segments are retained.

Each trajectory records joint positions, joint velocities, end-effector poses (position and orientation), and gripper states at 50\,Hz, providing a rich and flexible supervision signal that can support a variety of action representations and control modes during training.

During training, this dataset serves as the primary corpus for Stage~I T2A pretraining (\cref{sec:training-recipe}), where the DiT action decoder is initialized on language--action correspondences before visual grounding is introduced. By exposing the decoder to a large and diverse set of language--action pairs without visual conditioning, this component pre-empts visual shortcuts and establishes a strong action prior that transfers to and complements the downstream vision-conditioned training stages.

\subsubsection{Navigation data.}

We incorporate Navigation data (7.5\%), which stands out for its long-horizon trajectories coupled with rich visual information. The mobile robot is assumed to have 3 degrees of freedom: translation in the plane and rotation about the vertical axis (heading angle). Navigation data comprises navigation videos (sampled at 2 FPS), instructions, and trajectory information, jointly equipping the model with mobile capabilities such as instruction following, object searching, and target tracking.

\paragraph{Instruction following.} Instruction-following navigation data (4.3\%) covers a diverse range of navigation horizons and instruction granularities. In these episodes, the mobile robot must execute commands that intermix motion primitives (e.g., move forward, turn right/left, stop) with textual landmarks. This data encourages the model to align navigation history with the given instruction and predict the subsequent trajectory.

\paragraph{Object searching.} Object-searching data (2.3\%) derives from navigation histories where the robot seeks a text-described object within large-scale indoor environments. The robot is challenged to remember previously navigated areas while exploring efficiently to locate the target. Such episodes foster the model's long-horizon exploration and identification capabilities.

\paragraph{Target tracking.} Target-tracking data (1.0\%) consists of episodes in which the robot follows a moving target (e.g., a described person) through indoor spaces, maintaining a safe following distance and keeping the target in view. The resulting trajectories blend reactive motion adjustments with longer-horizon path decisions triggered by turns or occlusions.

\subsubsection{Vision-language data.}
% \subsubsection{Robotics Manipulation Trajectories}
% \subsubsection{Vision-Language Data}
\label{sec:vl-data}

We further incorporate auxiliary vision-language supervision (8.5\% combined)
to strengthen semantic grounding, fine-grained instruction following, and
general visual reasoning.

\paragraph{Fine-grained embodied action caption.}
Most existing robot datasets provide only coarse task-level labels
(e.g., ``pick up the ceramic bowl''), which are insufficient for precise
action prediction: the same label may correspond to vastly different
execution strategies (e.g., grasping from the left vs.\ the right,
rotating clockwise vs.\ counter-clockwise), creating ambiguity during
policy learning. Drawing inspiration from fine-grained text control in
video generation, we construct dense action descriptions that tightly
couple language to each individual episode.
To build a diverse yet manageable annotation set, we draw episodes from
major open-source manipulation datasets across all task categories and
apply a clustering-based selection strategy: episodes within each task
are grouped by their action information, and representative ones are
sampled to ensure broad coverage.
Each selected episode is annotated along 13 dimensions: action primitive,
actor identity, object recognition and disambiguation, contact region,
source and target location, trajectory and orientation, gripper state,
and body motion. We use a two-stage pipeline for annotation.
In the first stage, a vision-language model (Qwen3.6-plus) watches the
video and extracts a coarse action sequence along with the manipulated
object. In the second stage, the video is densely sampled into frames
to preserve subtle motion details; the model then re-watches these
frames conditioned on the first-stage output and produces fine-grained,
step-by-step descriptions enriched with contact points, spatial
relationships, and motion trajectories.
For multi-view datasets, an additional refinement pass incorporates
wrist- or side-camera footage to correct details only visible from
close-up or alternative perspectives.
All generated captions are finally reviewed and corrected by human
annotators.
This pipeline yields approximately 48{,}000 fine-grained video--caption
pairs (0.2\% of the pretraining mixture).
Table~\ref{tab:caption-example} contrasts a coarse label with our
fine-grained caption for the same episode.

\begin{table}[h]
\centering
\caption{Coarse task label vs.\ fine-grained action caption for the same episode.}
\label{tab:caption-example}
\setlength{\tabcolsep}{4pt}
\begin{tabular}{p{0.15\linewidth} p{0.75\linewidth}}
\toprule
\textbf{Coarse label} & ``Pick up, rotate, and place the ceramic bowl.'' \\
\midrule
\textbf{Fine-grained} &
  \textit{Step 1:} Pick up the ceramic bowl from the right far edge. \newline
  \textit{Step 2:} Rotate the bowl clockwise for two full circles. \newline
  \textit{Step 3:} Place the bowl at the center of the table. \\
\bottomrule
\end{tabular}
\end{table}

Building on the annotated data, we further fine-tune a dedicated VLM for
embodied action captioning. We also design a companion evaluation
benchmark consisting of QA pairs derived from the ground-truth
annotations together with a scoring rubric for assessing caption quality.

\paragraph{Autonomous driving VQA.}
We incorporate autonomous-driving visual question-answering data (2.4\%) as a complementary source of visually grounded, decision-oriented supervision. This mixture focuses on four transferable capabilities: (1) \emph{Temporal scene understanding}: driving scenes evolve under ego-motion, dynamic agents, and changing hazards. We include LingoQA~\citep{marcu2024lingoqa}, DriveAction~\citep{hao2025driveaction}, and MMAU~\citep{fang2024abductive} to strengthen event understanding and future-state prediction. 
(2) \emph{Surround-view spatial reasoning}: multi-camera observations provide wide-field geometric context. We include Impromptu-VLA~\citep{chiimpromptu}, nuScenes-QA~\citep{qian2024nuscenes}, nuScenes-MQA~\citep{inoue2024nuscenes}, MapLM~\citep{cao2024maplm}, and WaymoQA~\citep{yu2025waymoqa} to improve viewpoint-robust localization and spatial-relation understanding across diverse vehicles, sensors, and road layouts. 
(3) \emph{Language-grounded localization}: embodied agents must bind language to objects, regions, and spatial relations before acting. We include CODA-LM~\citep{chen2025automated}, Talk2Car~\citep{deruyttere2019talk2car}, DrivingVQA~\cite{corbiere2025retrieval}, DriveLM~\citep{sima2024drivelm}, W3DA~\citep{zhou2025towards}, and GRAID~\citep{elmaaroufi2025graid} for object-level grounding and region-aware question answering. 
(4) \emph{Planning-aware reasoning}: safe decision making requires linking perception to route, lane, and trajectory constraints. We include Bench2Drive-VL~\citep{jia2026bench2drive}, DriveGPT4~\citep{xu2024drivegpt4}, OmniDrive~\citep{wang2025omnidrive}, Senna~\citep{jiang2024senna}, NAVSIM-RecogDrive~\citep{li2025recogdrive}, and NAVSIM-Traj to provide supervision for maneuver selection and future trajectory prediction. 
Prior to pretraining, all data sources undergo standardization into a unified conversational format. We convert multiple-choice annotations into free-form QA supervision and normalize bounding boxes to a 1000-scale coordinate system. Additionally, multi-frame samples receive frame tags to maintain temporal order, while surround-view samples are assigned view tags to explicitly denote camera identity. Collectively, these structured sources offer transferable supervision for object-centric grounding, temporal state estimation, intent prediction, and goal-conditioned decision-making. This comprehensively complements the manipulation and navigation data utilized in our embodied pretraining.

\paragraph{Spatial grounding.}
We include 2D bounding box grounding data (2.5\%) to reinforce object-level
spatial understanding. Accurate spatial grounding is a prerequisite for
language-conditioned manipulation, where the model must localize task-relevant
objects from instruction-specified descriptions.

\paragraph{General vision-language data.}
To preserve robust visual perception and language grounding during embodied VLA pretraining, we supplement the primary action-centric dataset with a curated mixture of general vision-language data (3.4\% of total tokens). This auxiliary mixture explicitly comprises: (1) captioning data for image--text alignment; (2) knowledge-oriented VQA and visual reasoning for world knowledge retention; (3) OCR and text-grounding samples for fine-grained textual perception; (4) interleaved instruction-following data for stable cross-modal alignment; and (5) common grounding tasks (e.g., referring expression, spatial relation prediction) for object-level language grounding. 
Critically, to better support embodied reasoning, we strategically up-weight video-centric, spatial-relation, and 3D-aware supervision within this mixture. Collectively, these general vision-language objectives mitigate catastrophic forgetting, retain foundational world-knowledge priors, and enhance robustness in language-conditioned embodied perception.

\section{Post-Training}
\label{sec:post-training}

The large-scale pretraining of T2A and CPT produces \textbf{Qwen-VLA-Base}, a generalist vision-language-action model that exhibits broad cross-task and cross-embodiment generalization.
Although this broad coverage equips the model with versatile knowledge, it does not yet yield the precision needed for reliable closed-loop control on specific downstream tasks.
To bridge this gap, we introduce a two-stage post-training procedure that specializes base model for accurate task execution:
\begin{enumerate}[label=(\roman*),nosep]
    \item A multi-task \emph{supervised fine-tuning} (SFT) stage that jointly fine-tunes both the VLM backbone and the action expert on heterogeneous tasks, including visual question answering, spatial grounding, manipulation, and navigation, under embodiment-balanced and task-balanced sampling.
    \item A \emph{reinforcement learning} (RL) stage, initialized from the SFT checkpoint, that further refines the policy by optimizing directly against task-success-driven rewards obtained from on-policy rollouts in simulation, producing the final model \textbf{Qwen-VLA-Instruct}.
\end{enumerate}
Across both stages, we use cosine-decayed learning rates with separate group-wise schedules for the vision-language backbone and the action decoder, along with gradient clipping consistent with pretraining.
The specific data and methods for each stage are detailed in the subsequent sections.

\subsection{Multi-Task Supervised Fine-Tuning}
\label{sec:sft}

\paragraph{Data.}
The SFT data mixture comprises three categories: general vision-language samples, vision-language navigation episodes, and robotic manipulation demonstrations. 
First, we curate a set of vision-language samples---covering visual question answering, spatial grounding, and action captioning---to maintain the backbone’s visual understanding throughout fine-tuning. 
Second, we collect robotic manipulation demonstrations from simulation, including trajectories across a diverse set of platforms spanning single-arm, dual-arm, and humanoid robots, all using a prediction horizon of 16 action steps per chunk~\citep{community2026starvla}.
Following the same embodiment-aware prompt design used in pretraining (\Cref{sec:embodiment}), every training sample is prefixed with a textual prompt that specifies the robot platform, arm configuration, control frequency, and prediction horizon.
Third, we collect vision-language navigation episodes from continuous control environments, covering varied indoor scenes and instruction styles. We retain only successfully completed episodes, with a prediction horizon of 8 waypoints per chunk.

\paragraph{Objective.}
The SFT stage jointly optimizes two losses: next-token prediction on vision-language tokens and flow matching on action tokens. The loss weights are set to $0.1$ for vision-language next-token prediction and $1.0$ for both manipulation and navigation action prediction, ensuring that the model retains its language and visual understanding while focusing gradient capacity on action generation.

\subsection{Reinforcement Learning}
\label{sec:rl}

Likelihood-based SFT optimizes a surrogate that measures how well the policy \emph{imitates} demonstrations, but the quantity that ultimately matters in embodied control is closed-loop \emph{task success}: whether the executed trajectory achieves the goal.
These two objectives can diverge: a policy may assign high likelihood to plausible actions yet fail when its own outputs shift the state distribution away from the demonstration support.
The RL stage closes this gap by directly optimizing a task-success-driven reward on trajectories rolled out by the policy itself, producing the final model. We implement this stage using the RLinf framework~\citep{yu2025rlinf}.

\paragraph{Policy optimization objective.}
We adopt Proximal Policy Optimization (PPO)~\citep{ppo} with Generalized Advantage Estimation (GAE)~\citep{schulman2015high} as the RL algorithm.
Let $\pi_\theta$ denote the current policy and $\pi_{\theta_\text{old}}$ the policy that generated the rollout batch.
The clipped surrogate objective is
\begin{equation}
\label{eq:ppo}
\mathcal{L}^\text{actor}(\theta) = -\,\mathbb{E}_{t}\Big[\min\!\Big(
    r_t(\theta)\,\hat{A}_t,\;
    \text{clip}\!\big(r_t(\theta),\,1{-}\epsilon,\,1{+}\epsilon\big)\,\hat{A}_t
\Big)\Big],
\end{equation}
where $r_t(\theta) = \pi_\theta(a_t \mid s_t)\,/\,\pi_{\theta_\text{old}}(a_t \mid s_t)$ is the importance ratio, with $s_t = (o_t, x, e)$ the current state (visual observation, task instruction, and embodiment prompt following \cref{sec:Prompt-formulation}) and $a_t$ the predicted action chunk of horizon $H$, $\hat{A}_t$ is the GAE advantage estimate with discount $\gamma{=}0.99$ and trace-decay $\lambda{=}0.95$, and $\epsilon{=}0.2$ is the clipping threshold applied symmetrically.
The total loss combines the policy surrogate with a value-function regression term:
\begin{equation}
\label{eq:ppo-total}
\mathcal{L}(\theta) = \mathcal{L}^\text{actor}(\theta) + c_v\,\mathcal{L}^\text{value}(\theta),
\end{equation}
where $\mathcal{L}^\text{value}$ is a loss on value predictions and $c_v{=}1$.
We perform four optimization epochs per rollout batch.

\paragraph{Value estimation.}
Rather than training a separate critic network, we attach a lightweight value head directly to the vision-language backbone.
The value head mean-pools all VLM hidden states and maps them to a scalar value estimate through a linear projection.
We apply stop-gradient on the VLM hidden states before they enter the value head, so that value-function gradients do not propagate back through the pretrained backbone.
The value head is trained with a clipped MSE loss and a separate learning rate ($10^{-4}$, approximately $20\times$ the actor learning rate of $5{\times}10^{-6}$), allowing it to converge rapidly while the policy updates remain conservative.

\paragraph{Log-probability estimation under flow matching.}
A key challenge in applying PPO to our flow-matching action decoder is computing the log-probability $\log\pi_\theta(a_t \mid s_t)$ required for the importance ratio $r_t(\theta)$.
Unlike autoregressive token-based policies where log-probabilities are directly available from the softmax output, flow-matching models define an implicit density through a learned velocity field and iterative denoising.
We convert the deterministic probability-flow ODE into a corresponding SDE by injecting controlled noise at each Euler denoising step~\citep{song2021score}, so that each transition becomes an explicit Gaussian whose log-probability can be computed analytically without numerical ODE integration.
During rollout we store the intermediate denoising states; at the PPO update we re-evaluate the velocity field under the current parameters and recompute the Gaussian log-probability, yielding the importance ratio at negligible additional cost.
By default we randomly select a single denoising step per rollout for the log-probability estimate, requiring only one extra DiT forward pass during recomputation.
Both log-probabilities and advantages are computed at the \emph{action-chunk} level: each chunk of $H{=}16$ action steps receives a single scalar reward and a single advantage estimate, matching the temporal granularity of the flow-matching decoder's output.

\paragraph{Reward design.}
We use a sparse binary reward from the simulator: $R{=}1$ if the task goal is achieved at episode end and $R{=}0$ otherwise.
Credit assignment across action chunks is handled by GAE (\cref{eq:ppo}), which propagates the episode-level signal back through the value baseline.
No learned reward model is used; all signals come from the simulator's ground-truth task-completion semantics.

\paragraph{Rollout infrastructure.}
On-policy rollouts are collected in simulation environments that mirror the downstream evaluation benchmarks.
We adopt a decoupled client-server architecture in which a remote benchmark server hosts the simulation environments and the training process communicates with it via a network interface.
This separation allows the simulation workload to scale independently of the GPU-intensive policy optimization.
We instantiate $N{=}128$ parallel environment instances distributed across multiple manipulation task suites with non-uniform allocation proportional to task difficulty.
Each training iteration collects $8$ rollout epochs of $128$ environment steps each, yielding $N \times \lfloor 128/H \rfloor \times 8 = 128 \times 8 \times 8 = 8{,}192$ transition chunks per iteration (where $H{=}16$ is the action chunk length).
During rollout, actions are sampled from the policy with temperature $\tau{=}1.0$; at evaluation time, we reduce the temperature to $\tau{=}0.6$ to sharpen the action distribution.
The embodiment prompts used during rollout are identical to those in SFT, ensuring that the RL stage does not reintroduce a distribution shift in the prompt conditioning.

\paragraph{Out-of-domain generalization.}
Because the embodiment prompt is the sole platform-specific interface, the RL-refined policy transfers to out-of-domain environments, tasks, and embodiments without an additional adaptation head or domain-specific fine-tuning.
At deployment, we simply replace the embodiment prompt with the textual description of the physical robot platform, including its arm configuration, control frequency, and action space, while keeping the VLM backbone and DiT action decoder unchanged.
This zero-shot generalization is enabled by two properties of our training recipe: (1)~the CPT stage has already exposed the model to both simulated and real-world visual distributions (\cref{sec:training-recipe}), so the backbone's visual representations are not specific to the simulation renderer; and (2)~the RL stage optimizes task-success metrics that are invariant to the visual domain, encouraging the policy to rely on semantically meaningful visual features rather than renderer-specific textures.
We quantify the cumulative effect of each post-training stage in \cref{sec:ablation_rl}.

\section{Experiments}

We conduct extensive experiments to evaluate the performance of Qwen-VLA in both simulation and real-world settings, covering two core embodied AI domains: robotic manipulation and visual navigation. We evaluate two model variants throughout: \textbf{Qwen-VLA-Base}, trained with large-scale pretraining on diverse embodied data, and \textbf{Qwen-VLA-Instruct}, further fine-tuned with instruction-following data collected across diverse simulated environments and tasks, capable of executing fine-grained manipulation and navigation with greater precision and reliability in one unified model.

\subsection{Main Results}

\subsubsection{Manipulation Results in Simulation}

To comprehensively evaluate the manipulation capabilities of Qwen-VLA, we benchmark on four simulation environments covering single-arm and dual-arm settings: LIBERO~\citep{liu2023libero}, Simpler~\citep{li2024simpler}, RoboCasa-GR1 Tabletop Tasks~\citep{nasiriany2024robocasa}, and RoboTwin 2.0~\citep{mu2025robotwin}.

\begin{itemize}[leftmargin=*,itemsep=2pt]
\item \textbf{LIBERO} is a single-arm tabletop benchmark comprising diverse tasks across four splits: Libero-Spatial, Libero-Object, Libero-Goal, and Libero-Long.

\item \textbf{Simpler-WidowX} is a real-to-sim evaluation suite using the WidowX robot arm. 

\item \textbf{RoboCasa-GR1} is a bimanual humanoid benchmark with 24 atomic kitchen tasks. 

\item \textbf{RoboTwin 2.0} is a dual-arm benchmark with 50 bimanual tasks divided into Easy and Hard difficulty tiers.
\end{itemize}
Together, these benchmarks test a wide range of manipulation skills from tabletop pick-and-place to long-horizon bimanual kitchen tasks. For all benchmarks, we set the action chunk length $H=16$ and report the average success rate following the standard evaluation protocol in StarVLA~\citep{community2026starvla}.

\begin{table}[htbp]
  \centering
  \small
   \caption{\textbf{Robot manipulation results across benchmarks: specialists vs.\ a single generalist.} Specialist models are fine-tuned separately on each benchmark, whereas our generalist (Qwen-VLA) is trained once on all embodiments jointly and evaluated across all platforms without per-benchmark adaptation. \textbf{Bold}: best; \underline{underline}: second best.}
  \label{tab:sub_manip}
  \resizebox{\textwidth}{!}{%
  \begin{tabular}{llccccc}
    \toprule
    \multirow{2}{*}{\textbf{Method}} & \multirow{2}{*}{\textbf{Type}} &
      \multicolumn{5}{c}{\textbf{Benchmark}} \\
    & & \textbf{LIBERO} & \textbf{RoboCasa-GR1} & \textbf{Simpler-WidowX} & \textbf{RoboTwin-Easy} & \textbf{RoboTwin-Hard} \\
    \midrule
    $\pi_0$~\citep{black2024pi0}
      & \multirow{6}{*}{\textbf{Specialist}} & 94.4 & -- & -- & 65.9 & 58.4 \\
    StarVLA-OFT~\citep{community2026starvla}
      & & 96.6  & 48.8 & \underline{64.6} & 50.4 & -- \\
    GR00T N1.6~\citep{nvidia2025gr00t}
      & & 97.2 & 49.9  &  63.2  & 47.6 & -- \\
    $\pi_{0.5}$~\citep{black2025pi05}
      & & 97.6 & 37.0 & 46.9 & 82.7 & 76.8 \\
    ABot-M0~\citep{yang2026abotm0}
      & & \textbf{98.6} & \textbf{58.3} & -- & \underline{86.0} & \underline{85.0} \\
    Being-H0.5~\citep{luo2026beingh05}
      &  & 97.6 & 53.3 & -- & -- & -- \\
    \midrule
    \textbf{Qwen-VLA-Base}
      & \multirow{2}{*}{\textbf{Generalist}} & 90.8 & 40.4 & 64.3 & 64.3 & 66.4 \\
    \textbf{Qwen-VLA-Instruct}
      &  & \underline{97.9} & \underline{56.7} & \textbf{73.7}  & \textbf{86.1} & \textbf{87.2} \\
    \bottomrule
  \end{tabular}%
  }
\end{table}

Table~\ref{tab:sub_manip} compares our generalist model against state-of-the-art specialist policies. Specialist models are fine-tuned separately on each benchmark and remain embodiment-specific, while our single generalist is trained jointly on diverse, large-scale data spanning multiple scenes and embodiments, enabling deployment across all four platforms through embodiment-aware prompting alone. 

\paragraph{A single generalist outperforms most specialists.} Despite being an all-in-one model, Qwen-VLA-Instruct surpasses the majority of specialist baselines. On LIBERO, it achieves 97.9\%, on
par with the best specialists. On RoboCasa-GR1, it reaches 56.7\%, outperforming $\pi_{0.5}$ (37.0\%), GR00T N1.6 (49.9\%), and Being-H0.5 (53.3\%). On Simpler-WidowX, it attains
73.7\%, surpassing specialist baselines including StarVLA-OFT (64.6\%). On RoboTwin-Easy/Hard, it scores 86.1\%/87.2\%, \textbf{exceeding} the previous best specialist ABot-M0
(86.0\%/85.0\%). These results confirm that joint multi-embodiment training does not sacrifice task-specific performance; in several benchmarks, the generalist even outperforms dedicated
models.

\paragraph{Pretraining provides a strong foundation, and instruction tuning yields substantial gains.} Qwen-VLA-Base already achieves reasonable performance (e.g., 90.8\% on LIBERO, 64.3\% on
Simpler-WidowX), indicating that large-scale pretraining learns transferable manipulation primitives across embodiments. With instruction tuning, Qwen-VLA-Instruct improves consistently across
all benchmarks (+7.1\% on LIBERO, +16.3\% on RoboCasa-GR1, +9.4\% on Simpler-WidowX, +21.8\% on RoboTwin-Easy, +20.8\% on RoboTwin-Hard), confirming that a modest amount of task-specific alignment
is sufficient to convert general pretrained representations into precise, deployable policies.

\subsubsection{Manipulation Results in the Real World}
We collect the real-world dataset on ALOHA, a representative bimanual robot embodiment. The platform consists of two 6-DoF robotic arms equipped with parallel-jaw grippers, forming a bimanual manipulation system. It is instrumented with three RGB cameras, including two wrist-mounted cameras and one first-person-view camera.
Unlike Qwen-VLA-Instruct, which is fine-tuned on simulated environments, we further validate Qwen-VLA on real-world tasks by fine-tuning on demonstrations collected through ALOHA across multiple tasks. We prepare two variants sharing the same architecture: ${\text{Qwen-VLA-aloha}}_{\text{w/o pretrain}}$ trained from scratch, and ${\text{Qwen-VLA-aloha}}_{\text{w/ pretrain}}$ fine-tuned from Qwen-VLA-Base, to evaluate the transferability of large-scale embodied pretraining to real-world settings.

\begin{figure}[t]
\centering
\includegraphics[width=\linewidth]{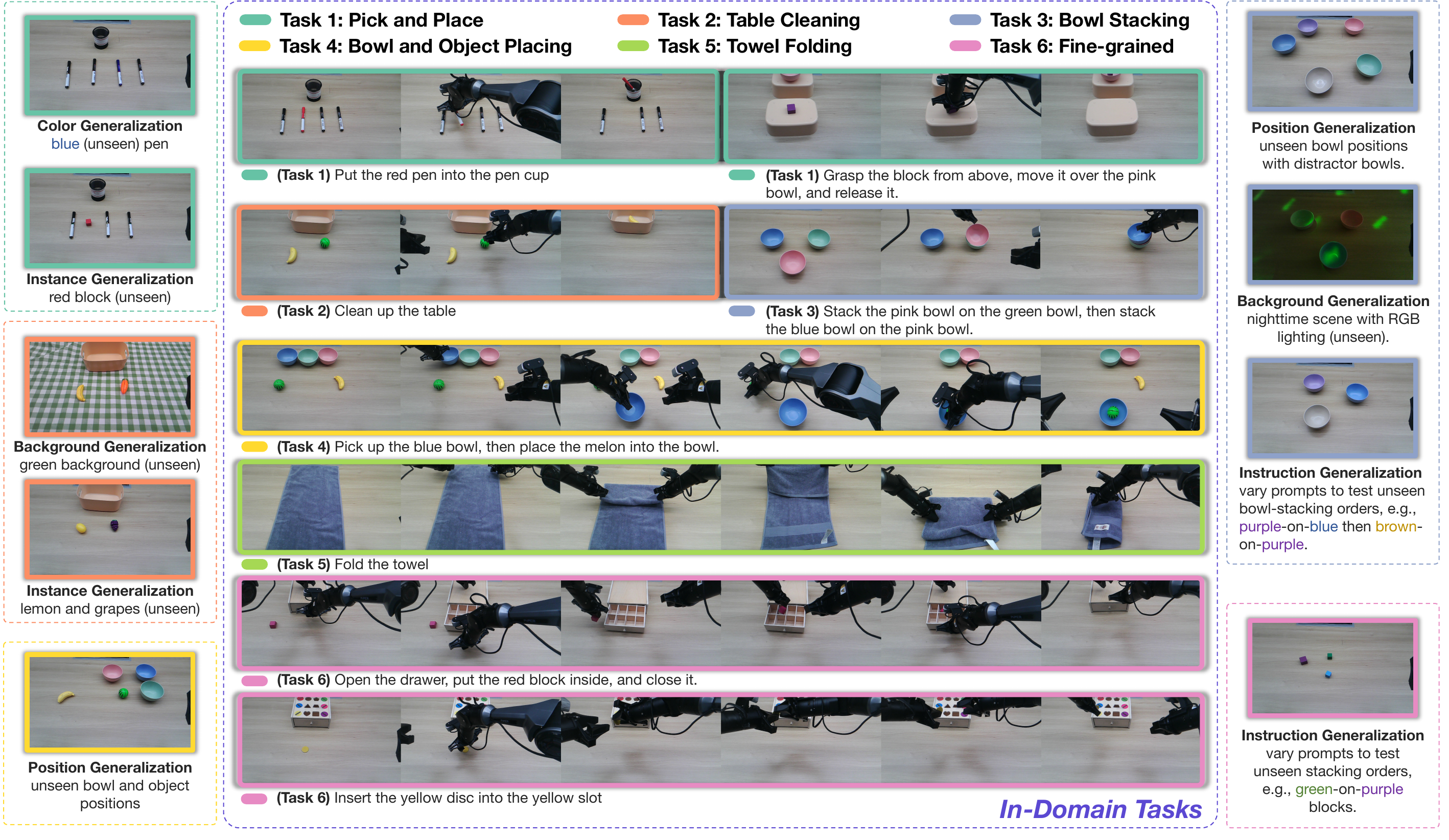}
% \caption{Task Overview.}
\caption{Overview of real-world evaluation tasks on the ALOHA bimanual platform.}
\label{fig:task_overview}
\end{figure}
We evaluate the model on a diverse set of real-world manipulation tasks, covering both in-domain and out-of-distribution (OOD) scenarios. The in-domain evaluation contains six task categories: pick and place, table cleaning, bowl stacking, bowl picking and object placement, towel folding, fine-grained manipulation. The OOD evaluation further tests generalization to unseen colors, instances, positions, background, language instructions.

\paragraph{In-domain tasks.} The in-domain evaluation covers six task categories of increasing complexity:
\begin{itemize}[leftmargin=*,itemsep=2pt]
    \item \textbf{Task 1: Pick and Place.} The robot picks up basic objects, including a block and a pen, and places them at specified target locations.
    
    \item \textbf{Task 2: Table Cleaning.} The robot cleans up objects from the table and places them into a target container.
    
    \item \textbf{Task 3: Bowl Stacking.} The robot stacks bowls in the color order specified by the language instruction.
    
    \item \textbf{Task 4: Bowl Pick \& Place.} The robot picks up a bowl and places target objects into it according to the language instruction.
    
    \item \textbf{Task 5: Towel Folding.} The robot folds a towel on the table, testing its ability to manipulate deformable objects.
    
    \item \textbf{Task 6: Fine-grained Manipulation.} This category includes several precise manipulation tasks, such as opening a drawer to place an object inside and then closing it, inserting an object into a slot, and stacking two blocks.
\end{itemize}

\paragraph{OOD tasks.}
The OOD evaluation includes five settings that assess different types of generalization:
\begin{itemize}[leftmargin=*,itemsep=2pt]
    \item \textbf{Color Generalization.} This setting evaluates color generalization on Task~1 and Task~4, where the robot needs to handle unseen color specifications, such as picking up a bowl with a specified color (unseen) and placing an object into it, or putting a pen with a specified color (unseen) into the target container.
    
    \item \textbf{Instance Generalization.} This setting evaluates instance generalization on Task~1 and Task~2, where the robot encounters novel objects, such as grapes (unseen) and a lemon (unseen) during table cleaning, or a red block (unseen) during object placement.
    
    \item \textbf{Position Generalization.} This setting evaluates position generalization on Task~3 and Task~4, where bowls and target objects appear at spatial configurations (unseen).
    
    \item \textbf{Background Generalization.} This setting evaluates background and lighting generalization on Task~2 and Task~3, including lighting conditions (unseen) for bowl stacking and a green background (unseen) for table cleaning.
    
    \item \textbf{Instruction Generalization.} This setting evaluates instruction generalization on Task~3 and Task~6, where the robot is required to execute different actions according to the given language instruction. Examples include stacking bowls in a specified order (unseen) or stacking blocks according to color relations, e.g., placing the green block (unseen) on the purple block (unseen).
\end{itemize}
\begin{table}[htbp]
    \centering
    \caption{\textbf{In-domain performance across short-horizon and long-horizon task categories.} ${\text{Qwen-VLA-aloha}}_{\text{w/o pretrain}}$ and ${\text{Qwen-VLA-aloha}}_{\text{w/ pretrain}}$ share the same Qwen-VLA model architecture; the former is trained from scratch, and the latter is fine-tuned from Qwen-VLA-Base.}
    \label{tab:indomain_results}
    \resizebox{\linewidth}{!}{%
    \begin{tabular}{lccccccc}
    \toprule
    \multirow{2}{*}{Model}
    & \multicolumn{3}{c}{Short-horizon Tasks}
    & \multicolumn{3}{c}{Long-horizon Tasks}
    & \multirow{2}{*}{Avg.} \\
    \cmidrule(lr){2-4} \cmidrule(lr){5-7}
    & Pick and Place
    & Table Cleaning
    & Bowl Stacking
    & Bowl Pick \& Place
    & Towel Folding
    & Fine-grained Manipulation
    & \\
    \midrule
    GR00T N1.6~\citep{nvidia2025gr00t}        & 30.8 & 38.5 & 53.8 & 19.2 & 19.2 & 10.3 & 28.6 \\
    $\pi_{0.5}$~\citep{black2025pi05}        & 73.1 & 84.6 & 88.5 & 69.2 & \textbf{80.8} & 33.3 & 71.6 \\
    ${\text{Qwen-VLA-aloha}}_{\text{w/o pretrain}}$  & 30.8 & 53.8 & 61.5 & 64.1 & 50.0 & 30.8 & 48.5 \\
    ${\text{Qwen-VLA-aloha}}_{\text{w/ pretrain}}$ & \textbf{96.2} & \textbf{92.3} & \textbf{98.7} & \textbf{87.2} & 65.4 & \textbf{61.5} & \textbf{83.6} \\
    \bottomrule
    \end{tabular}%
    }
    \end{table}
    
    \begin{table}[t]
    \centering
    \caption{\textbf{OOD performance across generalization categories.} ${\text{Qwen-VLA-aloha}}_{\text{w/o pretrain}}$ and ${\text{Qwen-VLA-aloha}}_{\text{w/ pretrain}}$ share the same Qwen-VLA model architecture; the former is trained from scratch, and the latter is fine-tuned from pretrained Qwen-VLA-Base.}
    \label{tab:ood_results}
    \resizebox{0.78\linewidth}{!}{%
    \begin{tabular}{lcccccc}
    \toprule
    Model & Color & Instance & Position &  Background & Instruction & Avg. \\
    \midrule
    GR00T N1.6~\citep{nvidia2025gr00t}        & 46.2 & 38.5 & 3.8  & 19.2 & 19.2 & 25.4 \\
    $\pi_{0.5}$~\citep{black2025pi05}        & 57.7 & 61.5 & 19.2 & 26.9 & 42.3 & 41.5 \\
    ${\text{Qwen-VLA-aloha}}_{\text{w/o pretrain}}$ & 42.3 & 30.8 & 34.6 & 30.8 & 42.3 & 36.2 \\
    ${\text{Qwen-VLA-aloha}}_{\text{w/ pretrain}}$          & \textbf{88.5} & \textbf{76.9} & \textbf{53.8} & \textbf{80.8} & \textbf{84.6} & \textbf{76.9} \\
    \bottomrule
    \end{tabular}%
    }
    \end{table}
    
    \paragraph{Results on in-domain tasks.}
        Table~\ref{tab:indomain_results} summarises in-domain results.
    ${\text{Qwen-VLA-aloha}}_{\text{w/ pretrain}}$ achieves the best average success rate of 83.6\%, outperforming strong baselines including GR00T N1.6~\citep{nvidia2025gr00t} and $\pi_{0.5}$~\citep{black2025pi05} on most task categories.
    The benefit of large-scale pretraining is evident from the comparison with
    ${\text{Qwen-VLA-aloha}}_{\text{w/o pretrain}}$:
    fine-tuning from Qwen-VLA-Base raises the average success rate from 48.5\% to 83.6\%, with particularly large gains on Pick and Place, Table Cleaning, Bowl Stacking, Bowl Pick \& Place, and Fine-grained Manipulation.
    These results confirm that pretraining provides a strong foundation for real-world manipulation that transfers effectively across diverse in-domain tasks.

    \paragraph{Results on OOD tasks.}
    Table~\ref{tab:ood_results} reports the OOD generalization performance. ${\text{Qwen-VLA-aloha}}_{\text{w/ pretrain}}$ consistently achieves the best results across all five generalization settings, including color, instance, position, background, and instruction generalization. Its average OOD success rate reaches 76.9\%, substantially outperforming $\pi_{0.5}$~\citep{black2025pi05} by 35.4 percentage points and ${\text{Qwen-VLA-aloha}}_{\text{w/o pretrain}}$ by 40.7 percentage points. The improvement is particularly significant under background and instruction generalization, where ${\text{Qwen-VLA-aloha}}_{\text{w/ pretrain}}$ achieves 80.8\% and 84.6\%, respectively. These results suggest that pretraining not only improves in-domain performance but also provides stronger robustness to unseen visual conditions, object instances, positions, and instruction variations.
    
    \paragraph{Overall analysis.}
    Overall, the results demonstrate that pretraining plays a critical role in real-world manipulation learning. While ${\text{Qwen-VLA-aloha}}_{\text{w/o pretrain}}$ and ${\text{Qwen-VLA-aloha}}_{\text{w/ pretrain}}$ share the same architecture, training from scratch leads to much lower performance, especially under OOD settings. This comparison indicates that the performance gain does not simply come from model architecture, but mainly from the pretrained Qwen-VLA-Base model. Moreover, the strong OOD performance of ${\text{Qwen-VLA-aloha}}_{\text{w/ pretrain}}$ shows that pretrained vision-language-action representations can effectively support generalization across both visual variations and instruction variations.
    
\subsubsection{Navigation Results}

We evaluate the navigation capabilities of Qwen-VLA on the vision-and-language navigation in continuous environments (VLN-CE~\citep{krantz_vlnce_2020,ku2020room}). Specifically, we test Qwen-VLA Base and Instruct versions on the Val-Unseen split of the R2R and RxR benchmarks, and compare them with the open-source baselines. We use the default setting of VLN-CE and, in particular, implement a \texttt{sliding-window} waypoint action compatible with the predicted trajectory of Qwen-VLA.

\begin{table}[htbp]
  \centering
  \small
  \caption{\textbf{Comparison with open-source baselines on VLN-CE.} We compare the Qwen-VLA Base and Instruct versions with widely regarded open-source baselines on the Val-Unseen split of the R2R and RxR benchmarks. \textbf{Bold}: best; \underline{underline}: second best.}
  \label{tab:sub_vlnce}
  \setlength{\tabcolsep}{3.5pt}
  \begin{tabular}{lcccc|cccc}
    \toprule
    \multirow{2}{*}{\textbf{Method}} &
      \multicolumn{4}{c|}{\textbf{R2R Val-Unseen}} &
      \multicolumn{4}{c}{\textbf{RxR Val-Unseen}} \\
    & NE$\downarrow$ & OS$\uparrow$ & SR$\uparrow$ & SPL$\uparrow$
    & NE$\downarrow$ & SR$\uparrow$ & SPL$\uparrow$ & nDTW$\uparrow$ \\
    \midrule
    NaVid~\citep{zhang2024navid}                & 5.7 & 49.2 & 41.9 & 36.5 & \textbf{5.7} & 45.7 & 38.2 & -- \\
    Uni-NaVid~\citep{zhang2024uni}              & 5.6 & 53.3 & 47.0 & 42.7 & 6.2 & 48.7 & 40.9 & -- \\
    NaVILA~\citep{cheng2024navila}              & 5.2 & 62.5 & 54.0 & 49.0 & 6.8 & 49.3 & 44.0 & \underline{58.8} \\
    StreamVLN~\citep{wei2025streamvln}          & \textbf{5.0} & \underline{64.2} & \underline{56.9} & \textbf{51.9} & 6.2 & 52.9 & \underline{46.0} & \textbf{61.9} \\
    
    \midrule
    \textbf{Qwen-VLA-Base}                        & 5.2 & 61.7 & 53.8 & 49.4 & 6.4 & \underline{55.1} & 45.8 & 56.2 \\
    \textbf{Qwen-VLA-Instruct}                        & \underline{5.1} & \textbf{69.0} & \textbf{57.5} & \underline{51.2} & \underline{5.8} & \textbf{59.6} & \textbf{47.8} & 57.1 \\
    \bottomrule
  \end{tabular}
\end{table}

As shown in Table~\ref{tab:sub_vlnce}, Qwen-VLA-Instruct achieves the best performance on most metrics across both R2R and RxR benchmarks. On R2R Val-Unseen, it attains the highest Oracle Success rate (69.0) and Success Rate (57.5), surpassing StreamVLN by 4.8 and 0.6 points respectively, while maintaining competitive SPL. On the more challenging RxR Val-Unseen split, Qwen-VLA-Instruct leads in both SR (59.6) and SPL (47.8), outperforming all baselines by a notable margin. Joint training on VLA and VLN data maintains reasonable performance on both benchmarks.

\subsubsection{Out-of-Distribution Evaluation on Static Manipulation}
To test whether pre-training priors generalise beyond the fine-tuning
  distribution, we construct a controlled OOD benchmark where fine-tuning
  covers only simple pick-and-place yet evaluation demands unseen task types.
We develop SimplerEnv-OOD, a suite of 6 out-of-distribution tasks spanning 3 tabletop scenes built upon the SimplerEnv framework with a WidowX robotic arm. All models are fine-tuned solely on the Bridge training split, which contains only simple pick-and-place demonstrations with fixed object pairings. None of the OOD task instructions, spatial relations, or manipulation primitives appear in the training data. The six tasks are:

\begin{itemize}[leftmargin=*,itemsep=2pt]      
      \item \textbf{MoveAway:} Move the spoon farther from the cloth such that the final distance exceeds the initial distance. This tests \textit{positional generalization} through distance-increasing placement.
      
      \item \textbf{MoveRight:} Move the spoon to the right side of the cloth. This tests \textit{positional generalization} through directional placement with orthogonal-axis constraints.                     

      \item \textbf{PlaceNear:} Move the carrot to a position adjacent to but not on top of the plate. This tests \textit{positional generalization} through proximity-based placement.
      
      \item \textbf{PlaceRight:} Place the spoon onto the right half region of the cloth. This tests \textit{positional generalization} with region-constrained placement, requiring the policy to interpret a sub-region specifier (``right half'') and achieve both correct target contact and
   spatial precision within the designated area.            

      \item \textbf{PutFront:} Place the carrot in front of the plate. This tests \textit{positional generalization} along the front/back directional axis.

      \item \textbf{StackYellow:} Pick up the yellow block and stack it on the green block. The training data only contains stacking green on yellow; this reversed order tests \textit{visual generalization} by requiring color-based reasoning and generalization to novel object-color bindings not seen during training.
  \end{itemize}

As shown in Table~\ref{tab:sub_ood}, Qwen-VLA-Instruct achieves the highest average success rate of \textbf{32.0\%}, substantially outperforming $\pi_{0.5}$ (12.6\%) across most tasks. The
advantage is strongest on positional generalization: $\pi_{0.5}$ fails completely on MoveRight and PlaceNear, while Qwen-VLA-Instruct reaches 33.3\% and 39.6\%, respectively. Consistent
gains also appear on MoveAway (43.8\% vs.\ 26.1\%) and PlaceRight (47.9\% vs.\ 32.1\%), indicating stronger generalization to unseen spatial instructions.
On StackYellow (22.9\% vs.\ 4.2\%), where the training data only contains stacking green on yellow but the task requires the reversed order, Qwen-VLA-Instruct demonstrates stronger generalization
to novel color-object bindings.

\begin{table}[htbp]
    \centering
    \small
    \caption{\textbf{OOD generalization results on SimplerEnv-OOD.}
      \textbf{Bold}: best; \underline{underline}: second best.}
    \label{tab:sub_ood}
    \setlength{\tabcolsep}{0pt}
    \begin{tabular*}{\textwidth}{l@{\extracolsep{\fill}}cccccc|c}
      \toprule
      \multirow{2}{*}{\textbf{Method}} &
        \multicolumn{6}{c|}{\textbf{OOD Task}} &
        \multirow{2}{*}{\textbf{Avg.}} \\
      & MoveAway & MoveRight & PlaceNear & PlaceRight & PutFront & StackYellow \\
      \midrule
      $\pi_{0.5}$~\citep{black2025pi05}
        & 26.1 & 0.0 & 0.0  & 32.1 & \textbf{13.0} & 4.2 & 12.6 \\
      \midrule
      \textbf{Qwen-VLA-Base}
        & \underline{31.3} & \underline{31.6} & \underline{16.7} & \underline{47.1} & \underline{6.3} & \underline{18.8} & \underline{25.3} \\
      \textbf{Qwen-VLA-Instruct}
        & \textbf{43.8} & \textbf{33.3} & \textbf{39.6} & \textbf{47.9} & 4.2 & \textbf{22.9} & \textbf{32.0} \\
      \bottomrule
    \end{tabular*}
\end{table}

\subsubsection{Out-of-Distribution Evaluation on Dynamic Manipulation}

Deploying robots in complex environments requires mastering dynamic manipulation and necessitating adaptation to independent object motion~\citep{liang2026cook}. To systematically evaluate this critical capability, the recently introduced DOMINO benchmark~\citep{fang2026towards} pioneers a comprehensive evaluation paradigm. It innovatively introduces a hierarchical motion design and a continuous Manipulation Score to capture execution quality under unpredictable constraints. These unique features make DOMINO an exceptional stress test for determining whether a vision-language-action model possesses generalizable spatial-to-kinematic mapping capabilities rather than merely executing memorized static routines. We evaluated Qwen-VLA in a zero-shot setting across all 35 DOMINO suites and report the Success Rate (SR) alongside the Manipulation Score (MS).

As shown in Table~\ref{tab:domino_ood}, Qwen-VLA establishes the best overall performance on the DOMINO benchmark. Qwen-VLA-Instruct achieves the highest SR (26.6\%) and MS (39.5). In the zero-shot category, our model demonstrates a substantial advantage over standard vision-language-action architectures by surpassing both OpenVLA-OFT~\citep{kim2025fine} and $\pi_{0.5}$~\citep{black2025pi05} by over 19 percentage points in SR. Furthermore, it edges out the representative WAM-style method LingBot-VA~\citep{li2026causal} while demonstrating stronger continuous execution quality under moving-object dynamics. Crucially, Qwen-VLA outperforms the entire suite of baselines explicitly fine-tuned for dynamic manipulation. For example, the advanced method PUMA~\citep{fang2026towards} relies on DOMINO-specific fine-tuning and temporal motion inputs to maintain dynamic awareness. Despite lacking these tailored adaptations and relying solely on current-frame observations, Qwen-VLA-Instruct surpasses PUMA by 9.4 percentage points in SR and 4.5 points in MS. Our model achieves these impressive gains at inference using only current-frame observations, without any dynamic-manipulation fine-tuning. This result highlights the advantage of our unified action-and-trajectory pretraining strategy in learning transferable spatial-to-kinematic priors.

We attribute this out-of-distribution generalization to the combination of decisive action generation and broad embodied pretraining. The flow-matching action decoder produces coherent action chunks. This mechanism reduces hesitation and helps the policy act precisely within narrow temporal windows. Additionally, large-scale joint pretraining across manipulation, navigation, trajectory prediction, and vision-language data provides transferable priors over visual grounding, spatial reasoning, and continuous control. These robust priors enable Qwen-VLA to generalize effectively to moving-object manipulation despite the complete absence of dynamic training data.

\begin{table}[t]
    \centering
    \small
    \caption{\textbf{Comparison with state-of-the-art methods on DOMINO.} \textbf{Bold}: best; \underline{underline}: second best.}
    \label{tab:domino_ood}
    \setlength{\tabcolsep}{6pt}
    \begin{tabular}{lcc}
      \toprule
      \textbf{Method} & \textbf{SR (\%)$\uparrow$} & \textbf{MS$\uparrow$} \\
      \midrule
      \multicolumn{3}{l}{\textit{Fine-tuned on dynamic manipulation data}} \\
      \midrule
      OpenVLA~\citep{kim2024openvla} & 1.5  & 6.1 \\
      RDT-1B~\citep{liu2025rdt} & 5.3  & 17.7 \\
      $\pi_0$~\citep{black2024pi0} & 8.2  & 24.0 \\
      $\pi_{0.5}$~\citep{black2025pi05} & 9.6  & 26.2 \\
      InternVLA-M1~\citep{chen2025internvla} & 5.4  & 27.6 \\
      VLA-Adapter~\citep{wang2026vla} & 4.4  & 24.3 \\
      $\pi_0$-FAST~\citep{pertsch2025fast} & 3.5  & 20.9 \\
      OpenVLA-OFT~\citep{kim2025fine} & 9.1  & 24.1 \\
      StarVLA-OFT~\citep{community2026starvla} & 10.9 & 30.5 \\
      PUMA~\citep{fang2026towards} & 17.2 & 35.0 \\
      \midrule
      \multicolumn{3}{l}{\textit{Zero-shot to dynamic manipulation}} \\
      \midrule
      OpenVLA-OFT~\citep{kim2025fine} & 6.7  & 20.0 \\
      $\pi_{0.5}$~\citep{black2025pi05} & 7.5  & 20.4 \\
      LingBot-VLA w/ depth~\citep{wu2026pragmatic} & 11.8 & 26.7 \\
      LingBot-VA~\citep{li2026causal} & \underline{24.1} & 36.1 \\
      \midrule
      \textbf{Qwen-VLA-Base} & 21.1  & \underline{37.4} \\
      \textbf{Qwen-VLA-Instruct} &
      \textbf{26.6}    & \textbf{39.5} \\
      \bottomrule
    \end{tabular}
\end{table}

\subsubsection{Out-of-Distribution Generalization in Real World}
\begin{figure}[h]
  \centering
  \includegraphics[width=\linewidth]{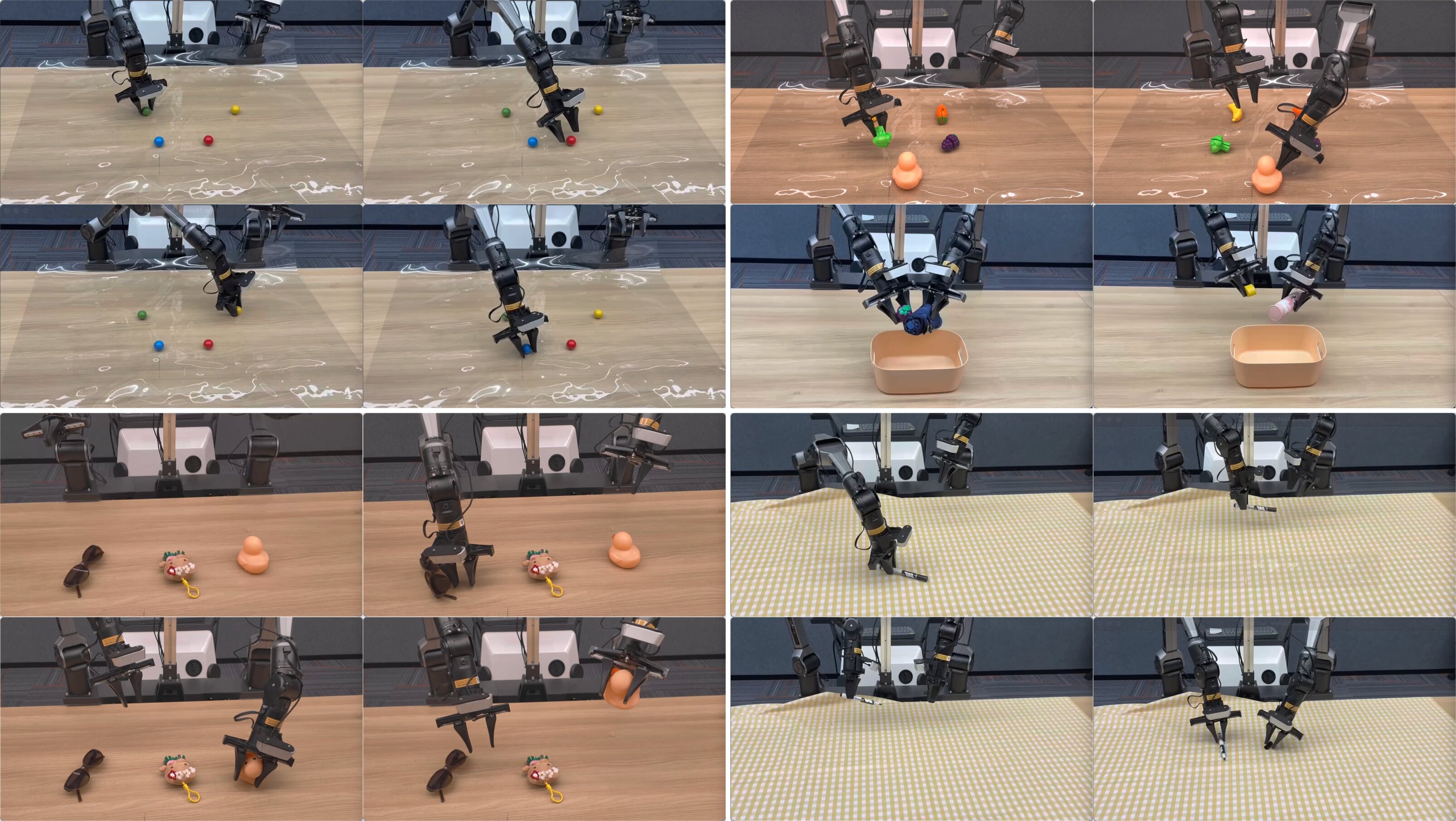}
  \caption{\textbf{Qualitative out-of-distribution generalization of Qwen-VLA-Base on the ALOHA dual-arm robot.}
  \textbf{Top-left:} color-conditioned grasping of green, blue, red, and yellow balls.
  \textbf{Top-right:} upper two panels show grasping of novel objects (green broccoli, toy duck);
    lower two panels show a compositional ``clean up the table'' task with sequential pick-and-place
    into a bin (blue umbrella, toy duck, bottled yogurt).
  \textbf{Bottom-left:} interaction with completely unseen objects (sunglasses, plush doll, toy duck).
  \textbf{Bottom-right:} atomic action transfer through uncapping a pen and placing the cap,
    robust to unseen yellow backgrounds.}
  \label{fig:qualitative_ood}
\end{figure}

To qualitatively examine continual pre-training, we evaluate \textbf{Qwen-VLA-Base} on the ALOHA dual-arm robot under four out-of-distribution settings: color grounding, novel object manipulation, unseen object recognition, and background robustness. Representative rollouts are shown in \cref{fig:qualitative_ood}.

\paragraph{Color grounding (top-left).}
We test whether Qwen-VLA-Base can discriminate between visually similar objects that differ only in color.
Across four trials the robot is prompted with ``grasp \{color\} ball'' for green, blue, red, and yellow, respectively.
In all cases the model correctly identifies the target color and produces accurate reaching-and-grasping trajectories, indicating color-conditioned semantic grounding after continual pre-training.

\paragraph{Novel object grasping and compositional clean-up (top-right).}
We next test whether the model can \emph{zero-shot} grasp objects never seen in manipulation pretraining data.
The top two panels show successful grasping prompted by ``grasp \{object\}'' for a green broccoli and a toy duck, both absent from the manipulation training set.
The model successfully localizes and grasps the broccoli; for the toy duck, it correctly identifies the object and reaches toward it, though the irregular shape prevents a stable grasp.
The bottom two panels further demonstrate a compositional task prompted by ``clean up the table,'' where the robot sequentially picks up a blue umbrella, a toy duck, and a bottled yogurt and places them into a bin. All three are everyday objects absent from the manipulation set.
Qwen-VLA-Base clears all objects in sequence, showing zero-shot visual recognition of novel objects and re-planning after each pick-and-place cycle.
This behavior is consistent with the mixed VL--VLA--VLN training setup: visual-language data exposes the model to a broader object vocabulary, which can transfer to manipulation even without in-domain demonstrations.

\paragraph{Unseen object interaction (bottom-left).}
We further probe zero-shot generalization by issuing ``approach \{object\}'' instructions for object categories entirely absent from pretraining: sunglasses, a plush doll, and a toy duck.
Notably, the ``approach'' action itself is almost never present in our manipulation data, yet the model correctly interprets the instruction and navigates the end-effector toward each target object.
For the toy duck, the model even succeeds in grasping it by the head.
For sunglasses, the model accurately localizes the object and approaches it, though the thin and flat geometry prevents a stable grasp. This reflects the physical boundary of shape generalization rather than a failure in visual recognition or instruction following.
This result suggests instruction-following transfer from mixed VL--VLA--VLN continual pre-training: the model can follow novel action verbs and recognize unseen objects using linguistic and visual coverage from the general-purpose pre-training data.

\paragraph{Background robustness (bottom-right).}
Finally, we probe robustness to visual domain shift.
Qwen-VLA-Base successfully uncaps a pen and places the cap on the table, which is a dexterous two-stage action requiring precise force control.
While this task may have been encountered during pre-training, the yellow background used here was never seen during training.
Despite this unseen visual context, the policy completes the task without any degradation.
This invariance to background perturbation is consistent with the visual diversity of the general-purpose VL understanding data used during continual pre-training. Exposure to natural images with varied scenes, lighting, and backgrounds may encourage the visual encoder to focus on task-relevant objects rather than spurious background cues.

\subsection{Ablation Studies}
\subsubsection{Text-to-Action Pretraining Ablations}
\label{sec:t2a_ablation}

As established in \cref{sec:training-recipe}, T2A trains the decoder to
reconstruct action distributions from language and embodiment prompts alone,
building a structured action prior before visual grounding begins.
We now ablate five design choices that shape this prior:
data composition, sequence prediction mode, visual input,
flow-matching timestep distribution, and training duration.
All ablations share the same downstream training recipe; results are
task-success rate (\%) after SFT from the T2A checkpoint on Simpler-WidowX.

\paragraph{Data composition.}
\label{sec:t2a_data_ratio}

T2A acquires action priors from language alone, so its
data composition directly shapes what the decoder learns.
We consider two sources.
\emph{Syn} (language-only synthetic trajectories from our simulation pipeline)
is cheap to scale and broad in task coverage, but its trajectories are
kinematically idealised.
\emph{Real} (real-robot teleoperation data, vision dropped) captures authentic
physical dynamics but is expensive to collect and narrower in task scope.

  \begin{figure}[!t]
      \centering
      \includegraphics[width=\linewidth]{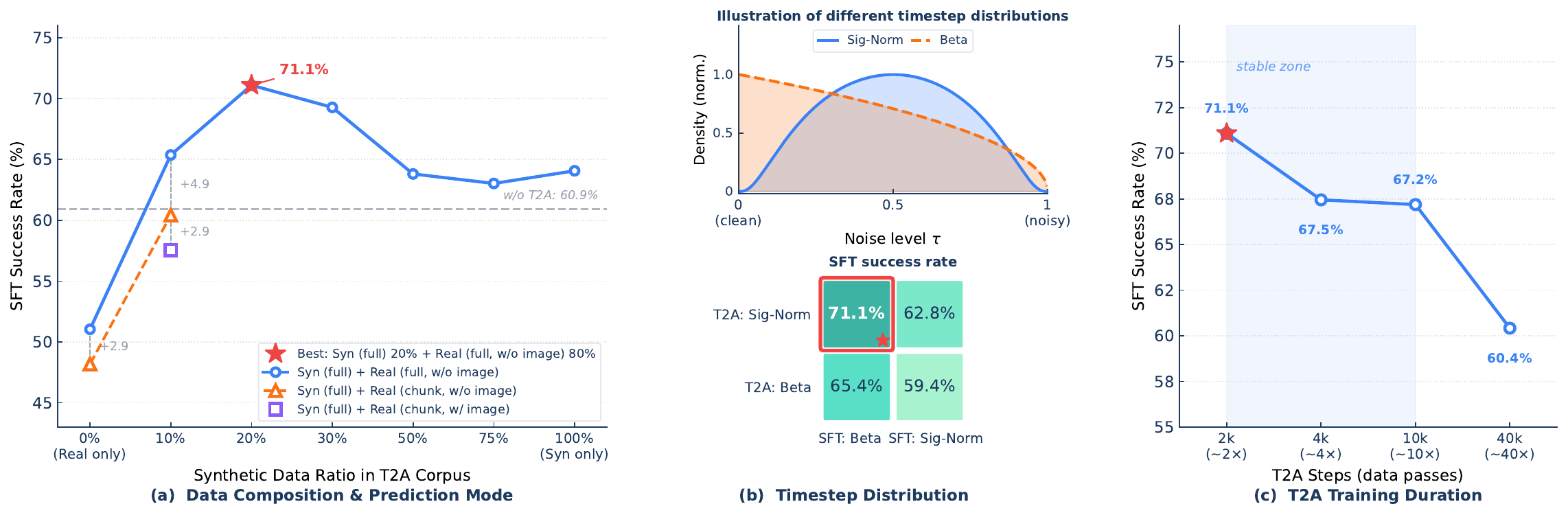}
      \caption{T2A pretraining ablations.
      \textbf{(a) Data composition and prediction mode.}
      SFT success rate (\%) vs.\ synthetic data ratio in the T2A corpus.
      The dashed line marks the baseline without T2A (60.9\%).
      Full-sequence prediction with $\sim$20\% synthetic + 80\% real data
      achieves the best result (71.1\%, $+$10.2~pp over no T2A).
      Chunk prediction consistently underperforms full-sequence (e.g.\
      $+$4.9 pp at 10\% synthetic), and including image tokens during T2A
      further hurts chunk-mode performance ($-$2.9 pp), confirming
      that images should be fully suppressed at T2A.
      \textbf{(b) Flow-matching timestep distribution.}
      The combination of Sigmoid-Normal $p(\tau)$ at T2A and Beta $p(\tau)$ at
      SFT achieves the highest success rate (71.1\%); reversing the choice
      at either stage degrades performance.
      \textbf{(c) T2A training duration.}
      Performance peaks at 2{,}000 steps (71.1\%) and plateaus through
      10{,}000 steps; 40{,}000 steps causes a slight drop due to
      overfitting to the T2A corpus.}
      \label{fig:t2a_ablations}
  \end{figure}

As shown in Fig.~\ref{fig:t2a_ablations}(a), purely real T2A data yields
51.0\%, while purely synthetic data reaches 64.1\%---better, but still
suboptimal.
Mixing $\sim$20\% synthetic with 80\% real achieves the best result
(\textbf{71.1\%}), a $+$10.2~pp gain over the no-T2A baseline (60.9\%).
% Note that simply freezing the VLM and skipping T2A does not suffice:
% the backbone's visual representations are optimised for perception, not
% for action generation, so aligning a randomly initialised decoder to
% those representations without first establishing an action prior leads
% to suboptimal convergence.
Syn broadens the coverage of language--action
correspondences that the decoder can learn, while Real anchors the prior
in physically plausible dynamics.

\paragraph{Sequence prediction mode: full-length vs.\ chunk.}

Full-sequence prediction supervises the entire trajectory in one forward pass,
exposing the decoder to long-horizon temporal structure and natural
start/termination patterns.
Chunk prediction instead splits each trajectory into fixed-length windows and
trains the decoder to produce only the next window from a task instruction,
which truncates long-horizon dependencies and creates positional ambiguity at
chunk boundaries.

Fig.~\ref{fig:t2a_ablations}(a) shows that full-sequence prediction
consistently outperforms chunk prediction across data mixes.
At 10\% synthetic data, the gap is $+$4.9~pp (65.4\% vs.\ 60.4\%);
at 0\% synthetic (real-only) the gap is $+$2.9~pp (51.0\% vs.\ 48.2\%).
Full trajectories let the decoder learn how language maps to complete action
sequences, including trajectory-level coherence and compositionality, whereas
chunks provide only local fragments that leave this mapping incomplete.
Importantly, the action representation is identical between T2A and
downstream stages: actions are encoded as delta end-effector displacements
relative to the first frame of each chunk, and the embodiment prompt
specifies the robot platform, normalisation convention, and prediction
horizon, allowing the decoder to handle heterogeneous embodiments within
the same T2A objective.

\paragraph{Visual input during T2A.}

T2A is deliberately a \emph{vision-free} stage.
Including image observations would raise per-step cost by roughly an order of
magnitude and, more importantly, would let the decoder shortcut through
visual cues instead of learning to ground actions in language.

Fig.~\ref{fig:t2a_ablations}(a) confirms this: at 10\% synthetic
data, chunk prediction with images scores 57.6\%, while the same setting
without images reaches 60.4\%---a $-$2.9~pp penalty for including vision.
When images are present, the decoder can rely on visual correlations
rather than building a reliable language--action mapping, diluting exactly
the prior that T2A is designed to establish.
{Image tokens should be fully suppressed during T2A}; visual grounding
is deferred to CPT, by which time the decoder already possesses a
well-formed action prior.

\paragraph{Flow-matching timestep distribution.}

Flow-matching policies condition on a scalar timestep $\tau \in [0,1]$ that
interpolates between the clean target ($\tau{=}0$) and pure noise ($\tau{=}1$).
The sampling distribution $p(\tau)$ determines which noise levels receive the
most gradient signal during training. The standard practice is to use a \emph{Beta}
distribution, which concentrates density toward the clean end; we follow this
convention at the CPT and SFT stages.
However, since T2A operates without visual conditioning, the decoder has no
backbone features to guide denoising, and we hypothesise that shifting
gradient toward intermediate timesteps---where the signal-to-noise ratio is
most informative for learning the language--action mapping---may be more
effective.
We therefore compare Beta against \emph{Sigmoid-Normal}, which peaks at
intermediate noise levels (Fig.~\ref{fig:t2a_ablations}(b)).

As shown in Fig.~\ref{fig:t2a_ablations}(b), Sigmoid-Normal $p(\tau)$ at T2A
combined with Beta $p(\tau)$ at SFT achieves the highest success rate
(\textbf{71.1\%}).
Swapping to Beta at T2A drops performance to 65.4\% ($-$5.7~pp), while
using Sigmoid-Normal at SFT instead of Beta yields 62.8\% ($-$8.3~pp).
The worst combination---Beta at both stages---reaches only 59.4\%.
Without visual conditioning to guide denoising, intermediate timesteps carry
the most learning signal for the language--action prior;
once the VLM backbone supplies rich conditioning at CPT/SFT, a Beta-shaped
$p(\tau)$ that spreads gradient more uniformly becomes more sample-efficient.
We therefore use {Sigmoid-Normal at T2A} and {Beta at CPT/SFT}
throughout our final pipeline.

\paragraph{T2A training duration.}

Longer T2A training lets the decoder more fully explore the action
distribution, but excessive training on a fixed T2A corpus risks overfitting
to its idiosyncrasies and reducing the plasticity available to CPT.
We sweep T2A duration over a wide range to locate the sweet spot.

As shown in Fig.~\ref{fig:t2a_ablations}(c), performance peaks at
2{,}000 steps (\textbf{71.1\%}) and remains comparable at 4{,}000 steps
(67.5\%) and 10{,}000 steps (67.2\%).
At 40{,}000 steps we observe a notable degradation to 60.4\%.
The rapid convergence is expected: the language--action mapping that T2A
learns is structurally lower-dimensional than the full
vision--language--action space, so a modest number of steps suffices to
capture it; prolonged training begins to memorise specific trajectory
instances rather than refining the structural prior.
We adopt \textbf{2{,}000 T2A steps} as the default in our final pipeline,
offering a favourable cost--quality tradeoff.

\subsubsection{Multi-Embodiment Co-training}
\label{sec:ablation_cotrain}

We study two design axes that are critical for multi-embodiment co-training: the role of vision-language data during action learning and the projection policy for heterogeneous action spaces.

\begin{figure}[!th]
      \centering
      \includegraphics[width=\linewidth]{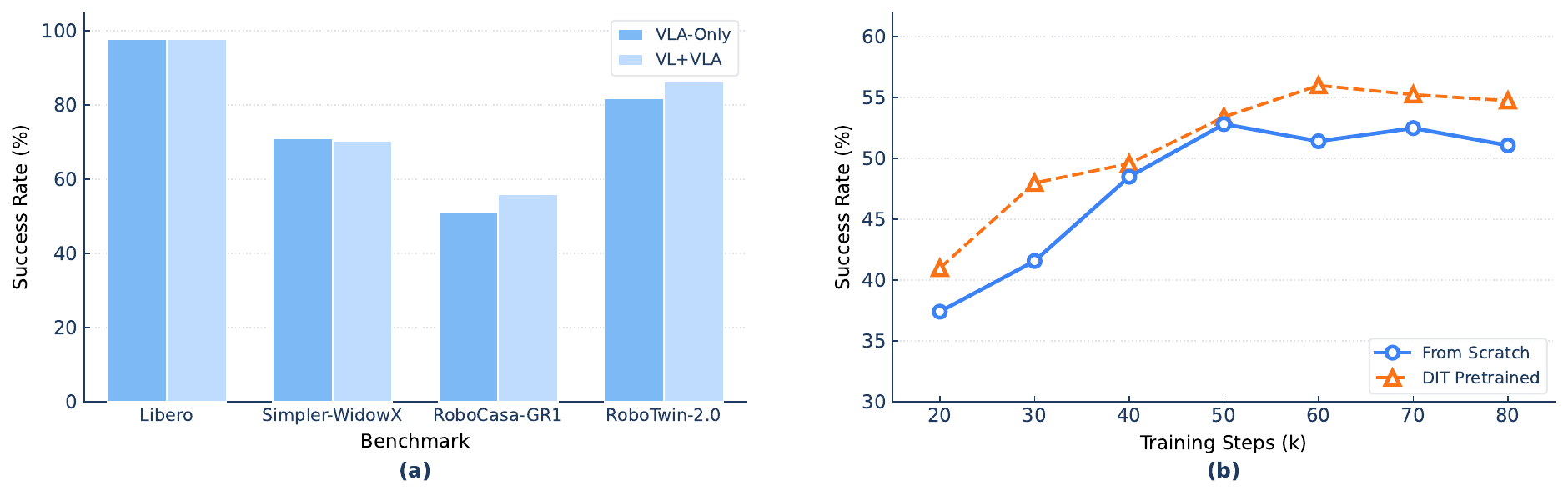}
      \caption{Vision-language co-training ablations.
      \textbf{(a) Impact of VL data on action learning.}
      Task-average success rate (\%) across four benchmarks for VLA-Only
      (action data only) vs.\ VL+VLA (action + vision-language data
      co-training).
      On simpler benchmarks (Libero, Simpler-WidowX) the two
      configurations perform comparably, confirming that VL co-training introduces no
      interference.
      On benchmarks that demand fine-grained object recognition and compositional instruction parsing, VL+VLA
      yields clear gains: +4.9~pp on RoboCasa-GR1 (51.1\% $\to$ 56.0\%)
      and +4.6~pp on RoboTwin-2.0 (81.8\% $\to$ 86.4\%).
      \textbf{(b) Transferability of pretrained DiT.}
      SFT success rate (\%) on RoboCasa-GR1 vs.\ training steps (k) when
      pairing the original Qwen3.5 VLM backbone with a DiT decoder
      initialised from scratch vs.\ a pretrained DiT.
      The pretrained DiT converges faster and reaches a higher peak.}
      \label{fig:vl_impact}
  \end{figure}

\paragraph{Impact of vision-language data on action learning.}
Although vision-language supervision operates in discrete token space
while action prediction targets continuous trajectories, the two
objectives share a common perceptual and linguistic backbone.
A natural question is whether retaining VL data during embodied
fine-tuning helps or hurts action quality.
We compare three configurations in Fig.~\ref{fig:vl_impact}:
\textbf{(a)}~\emph{Action-only}: the backbone is fine-tuned exclusively
on embodied trajectories with no VL supervision;
\textbf{(b)}~\emph{Action + VL data}: a fraction of VL data (VQA,
captioning, spatial grounding) is mixed into the training corpus.

As shown in Fig.~\ref{fig:vl_impact} (a), mixing VL data into training yields clear benefits on manipulation benchmarks.
On Libero and Simpler-WidowX, VLA-Only and VL+VLA achieve nearly identical success rates.
On RoboCasa-GR1, which requires whole-body control in cluttered kitchen scenes with diverse objects and spatial configurations, VL+VLA improves over VLA-Only by \textbf{+4.9\%} absolute (51.1\% $\to$ 56.0\%). 
Moreover, we attach the pretrained DiT to a fresh Qwen3.5-4B
VLM and compare against
a from-scratch baseline where the DiT is randomly initialised.
As shown in Fig.~\ref{fig:vl_impact} (b), the pretrained DiT
consistently outperforms the from-scratch baseline throughout training,
converging faster in early stages and reaching a higher peak success.

\paragraph{Projection design for heterogeneous embodiments.}

Different robot platforms have different action dimensionalities and
semantics (e.g.\ 7-DoF arm vs.\ 29-DoF whole-body).
To learn a shared representation space across heterogeneous action
formats, the projection between the DiT latent space and per-embodiment
action vectors must reconcile this mismatch.
To this end, we compare three representative designs.

\textit{Multi-MLP.}
Each embodiment owns a private encoder MLP that maps its native action
dimension to the DiT hidden size, and a private decoder MLP that maps
back. 
Let $d_i$ denote the action dimensionality of embodiment $i$, $h$ the
DiT hidden size, and $N$ the number of embodiments.  Each embodiment
owns a private encoder $f_i^{\mathrm{enc}}\!: \mathbb{R}^{d_i}\!\to\!\mathbb{R}^{h}$
and decoder $f_i^{\mathrm{dec}}\!: \mathbb{R}^{h}\!\to\!\mathbb{R}^{d_i}$,
totalling \textbf{$2h\!\sum_{i=1}^{N}\!d_i$}  parameters.

\textit{Concatenation.}
Actions from all $N$ embodiments are concatenated into a single vector
of dimensionality $D = \sum_{i=1}^{N} d_i$; each embodiment occupies a
dedicated, non-overlapping segment.  A shared encoder
$f^{\mathrm{enc}}\!: \mathbb{R}^{D}\!\to\!\mathbb{R}^{h}$ and decoder
$f^{\mathrm{dec}}\!: \mathbb{R}^{h}\!\to\!\mathbb{R}^{D}$ process the
full vector (\textbf{$2h\!\sum_{i=1}^{N}\!d_i$} parameters), and each embodiment retrieves its
predictions from the corresponding segment.

\textit{Zero-Padding.}
A single shared MLP encodes all embodiments; shorter action vectors are right-padded with zeros to
$d_{\max} = \max_i\, d_i$ before encoding.  A shared encoder
$f^{\mathrm{enc}}\!: \mathbb{R}^{d_{\max}}\!\to\!\mathbb{R}^{h}$ and
decoder $f^{\mathrm{dec}}\!: \mathbb{R}^{h}\!\to\!\mathbb{R}^{d_{\max}}$
process the padded vector (\textbf{$2h\,d_{\max}$} parameters); each embodiment
reads only its $d_i$-dimensional prefix.  Since
$d_{\max} \leq \sum_i d_i$, \textit{Zero-Padding} uses fewer projection
parameters than both \textit{Multi-MLP} and \textit{Concatenation}.

\begin{table}[htbp]
\centering
\caption{\textbf{Projection design ablation.} Results are task-average success
  rates (\%). All co-training variants share the same training recipe.}
\label{tab:projection_head}
\small
\begin{tabular}{l cc ccc}
\toprule
& \multicolumn{2}{c}{Single-Embodiment Training} & \multicolumn{3}{c}{Multi-Embodiment Co-training} \\
\cmidrule(lr){2-3} \cmidrule(lr){4-6}
& Bridge Only & Robocasa Only & Multi-MLP & Concat. & Zero-Pad \\
\midrule
Bridge   & 62.8 & ---  & \textbf{63.3} & 63.0 & 63.0 \\
Robocasa & ---  & 53.4 & 52.1 & 52.8 & \textbf{53.2} \\
\bottomrule
\end{tabular}
\end{table}

As shown in Table~\ref{tab:projection_head}, all co-training
projection designs achieve performance comparable to the
single-embodiment baselines on both benchmarks (Bridge single-train
62.8\% vs.\ co-train $\sim$63\%; Robocasa single-train 53.4\% vs.\
co-train 52--53\%), confirming that multi-embodiment co-training
introduces no significant penalty relative to training on each dataset
independently.
Among the three designs, performance differences are marginal
($<$1.2\% absolute on both benchmarks), indicating that the choice of
projection architecture has limited impact on task success once a shared
latent space is established.
However, Zero-Padding is architecturally the lightest: it requires only
$2h\,d_{\max}$ projection parameters compared with $2h\sum_i d_i$ for
Multi-MLP and Concatenation.
We therefore adopt \textbf{Zero-Padding} as the default projection
design.

\subsubsection{Effect of RL Post-Training}
\label{sec:ablation_rl}

We study whether RL on top of SFT provides additional gains, and whether those gains transfer to benchmarks not seen during RL training.
To isolate the contribution of each post-training stage, we evaluate three checkpoints along the training pipeline: the CPT base model (\textbf{Qwen-VLA-Base}), the model after multi-task SFT (\textbf{+SFT}), and the model after RL refinement (\textbf{+RL}, i.e., Qwen-VLA-Instruct).
RL rollouts are collected exclusively in SimplerEnv with sparse binary success rewards; no other benchmark data is used during the RL stage.

\begin{table}[htbp]
  \centering
  \small
  \caption{\textbf{Cumulative effect of post-training stages.} Each row adds one stage on top of the previous checkpoint. RL rollouts are collected only in SimplerEnv. All numbers are success rates (\%) except the DOMINO Manipulation Score (MS).}
  \label{tab:rl_effect}
  \resizebox{\textwidth}{!}{%
  \begin{tabular}{lcccccccc}
    \toprule
    \multirow{2}{*}{\textbf{Stage}} &
      \multicolumn{5}{c}{\textbf{In-Distribution}} &
      \multicolumn{1}{c}{\textbf{OOD-static}} &
      \multicolumn{2}{c}{\textbf{OOD-dynamic (DOMINO)}} \\
    \cmidrule(lr){2-6} \cmidrule(lr){7-7} \cmidrule(lr){8-9}
    & Simpler & RoboCasa & RoboTwin-E & RoboTwin-H & LIBERO
    & SimplerOOD & SR & MS \\
    \midrule
    CPT
      & 64.3 & 40.4 & 64.3 & 66.4 & 90.8
      & 25.3 & 21.1 & 37.4 \\
    \;+ SFT
      & 70.8 & 56.0 & 86.3 & 87.1 & 97.8
      & 31.6 & 25.7 & 39.1 \\
    \;+ RL
      & \textbf{73.7} & \textbf{56.7} & 86.1 & \textbf{87.2} & \textbf{97.9}
      & \textbf{32.0} & \textbf{26.6} & \textbf{39.5} \\
    \bottomrule
  \end{tabular}%
  }
\end{table}

As shown in Table~\ref{tab:rl_effect}, each post-training stage provides complementary gains.
SFT delivers large improvements across all benchmarks by specializing the general-purpose CPT representations to downstream task distributions.
RL then provides a further uplift on top of SFT, with the largest gain on SimplerEnv (+2.9~pp, from 70.8\% to 73.7\%), the environment in which RL rollouts are collected.
This confirms that directly optimizing closed-loop task success captures policy improvements that the imitation objective alone cannot.

More importantly, the RL gains are not confined to the training environment.
On benchmarks that are \emph{not} part of the RL rollout distribution, performance is preserved or mildly improved: RoboCasa increases by 0.7~pp, RoboTwin-Hard by 0.1~pp, LIBERO by 0.1~pp, and the OOD evaluation on SimplerEnv-OOD by 0.4~pp.
RoboTwin-Easy exhibits negligible changes ($<$0.6~pp), confirming that RL does not cause catastrophic forgetting on held-out tasks.
On the DOMINO dynamic manipulation benchmark, which tests zero-shot generalization to moving objects, both SR (25.7\% $\to$ 26.6\%) and MS (39.1 $\to$ 39.5) improve despite the complete absence of dynamic manipulation data during RL training.
These results indicate that task-success optimization in simulation preserves performance with mild positive transfer: the policy learns to execute more decisively and recover from compounding errors, benefits that generalize across evaluation settings.

\subsubsection{State Conditioning}
\label{sec:ablation_state}

Many robot learning methods condition the policy on explicit proprioceptive state (e.g., joint angles) in addition to visual observations, under the assumption that knowing the current robot configuration helps predict the next action.
However, proprioceptive inputs also introduce embodiment-specific dependencies that may hinder cross-embodiment generalization, since different platforms have different joint spaces and kinematic structures.
We examine whether explicit state conditioning benefits action learning in Qwen-VLA and, if so, where in the architecture the state should be injected.
We compare three strategies on RoboTwin-2.0:

\begin{itemize}[leftmargin=*,itemsep=2pt]
\item \textbf{No State.} The model receives only visual observations and the language instruction. No proprioceptive information is provided to either the VLM backbone or the DiT action decoder.

\item \textbf{State in VLM Prompt.} Joint angles are discretized into 256 bins per dimension and serialized into the language prompt, e.g., \texttt{"state: [34, 128, 67, 200, 55, 12, 88] pick up the bowl"}. The VLM processes state tokens together with images and the instruction; the DiT receives no additional state input.

\item \textbf{State in DiT.} Joint angles are passed as continuous-valued vectors directly to the DiT action decoder as an additional conditioning input, bypassing the VLM entirely.
\end{itemize}

\begin{table}[!th]
\centering
\caption{State conditioning ablation on RoboTwin-2.0 (success rates, \%).}
\label{tab:state_conditioning}
\small
\begin{tabular}{lcc}
\toprule
Conditioning & RoboTwin-Easy & RoboTwin-Hard \\
\midrule
No State & 88.7 & 87.4 \\
State in VLM Prompt & 89.3 & 88.7 \\
State in DiT & 89.4 & 88.3 \\
\bottomrule
\end{tabular}
\end{table}

As shown in Table~\ref{tab:state_conditioning}, incorporating proprioceptive state yields only marginal gains regardless of the injection strategy: at most +0.7~pp on Easy and +1.3~pp on Hard relative to the vision-only baseline.
The two injection strategies perform comparably ($\leq$0.4~pp difference), suggesting that neither the VLM's semantic processing nor the DiT's continuous conditioning offers a clear advantage for encoding joint-angle information.
We attribute the limited impact of state conditioning to two factors.
First, the multi-view visual observations already provide sufficient information about the robot's current configuration, especially for tasks where the end-effector is visible in at least one camera view.
Second, the flow-matching action decoder predicts relative action displacements rather than absolute poses, reducing the need for an explicit current-state reference.
Given the limited benefits and the additional complexity of maintaining embodiment-specific state interfaces across diverse platforms, we opt not to include state conditioning in our default framework, keeping the embodiment-aware text prompt as the sole platform-specific input.

\section{Conclusion}

We presented \textbf{Qwen-VLA}, a unified vision-language-action model that extends the Qwen vision-language backbone from perception and reasoning to embodied action generation. By formulating manipulation, navigation, egocentric action modeling, and trajectory prediction within a shared action-and-trajectory prediction framework, Qwen-VLA enables a single model to learn from heterogeneous embodied data across tasks, environments, and robot embodiments. Through embodiment-aware prompt conditioning, a DiT-based flow-matching action decoder, large-scale joint pretraining, and SFT/RL post-training, Qwen-VLA achieves strong generalist performance while retaining flexibility across different control conventions. Our results suggest that language-grounded VLA models provide a practical path toward embodied intelligence by aligning multimodal perception, instruction understanding, embodiment constraints, and executable actions. Compared with visual prediction-centric world models, Qwen-VLA emphasizes grounding multimodal understanding into actions that can be executed by embodied agents. Together, these results indicate that VLA models can serve as an actionable interface between multimodal foundation models and embodied agents, bridging high-level language reasoning with low-level physical control.

\section{Limitations and Future Work}

Despite its generalist capabilities, Qwen-VLA still has several limitations. First, embodied action data remains far smaller and less diverse than vision-language pretraining data, limiting robustness to long-tail objects, environments, embodiments, and contact-rich interactions. Second, joint training across vision-language understanding, navigation, and action generation introduces optimization trade-offs. While action-oriented training improves policy learning, it can modestly regress some pure vision-language and navigation evaluations, suggesting the need for better objective balancing, data curricula, and modular specialization. Third, current evaluations are still largely short-horizon and benchmark-driven, leaving long-duration, failure-prone real-world deployment as an open challenge.

These limitations motivate several future directions. Scaling real-world interaction data through autonomous collection, simulation, and sim-to-real transfer will be critical for improving robustness. Large-scale human video, from both egocentric and third-person views, can provide physical priors and temporal abstractions beyond robot demonstrations. Future models should incorporate long-horizon planning, episodic memory, and world modeling, enabling agents to track state, decompose tasks, recover from failures, and anticipate action consequences. Finally, richer physical feedback such as force, tactile, and proprioceptive signals, combined with large-scale reinforcement learning in simulation and the real world, may further bridge the gap between language-grounded embodied understanding and reliable control.

\section{Contributions and Acknowledgments}

\subsection*{Core Contributors}

\noindent
\begin{tabularx}{\textwidth}{@{}LLLLL@{}}
Qiuyue Wang$^{*}$ & Mingsheng Li$^{*}$ & Jian Guan$^{*}$ & Jinhui Ye & Sicheng Xie \\
Yitao Liu & Junhao Chen & Zhixuan Liang & Jie Zhang & Xintong Hu \\
Xuhong Huang & Pei Lin & Junyang Lin & Dayiheng Liu & Shuai Bai$^{\dagger}$ \\
Jingren Zhou & & & & \\
\end{tabularx}

\vspace{1.2em}

\subsection*{Contributors}

\noindent
\begin{tabularx}{\textwidth}{@{}LLLLL@{}}
Jiazhao Zhang & Haoqi Yuan & Gengze Zhou & Hang Yin & Ye Wang \\
Yiyang Huang & Zixing Lei & Wujian Peng & Delin Chen & Yingming Zheng \\
Jingyang Fan & Xianwei Zhuang & Xin Zhou & Haoyang Li & Anzhe Chen \\
Tong Zhang & Xuejing Liu & Yuchong Sun & Ruizhe Chen & Zhaohai Li \\
Chenxu L\"u & Zhibo Yang & Tao Yu & Xionghui Chen & \\
\end{tabularx}

\vspace{1.2em}

\subsection*{Acknowledgements}

We thank Jun Huang, Jianlou Si, Chao Wang, and Nana Jiang for their strong support on simulation environments and reinforcement learning. 

\vfill

\noindent
{\footnotesize
$^{*}$Equal contribution. \quad
$^{\dagger}$Corresponding author.
}

\bibliography{colm2024_conference}
\bibliographystyle{colm2024_conference}

\appendix

\end{document}